\documentclass[10pt]{article} 
\usepackage[preprint]{tmlr}

\usepackage{amsmath,amsfonts,bm}

\def\eqref#1{equation~\ref{#1}}

\def\1{\bm{1}}

\DeclareMathAlphabet{\mathsfit}{\encodingdefault}{\sfdefault}{m}{sl}
\SetMathAlphabet{\mathsfit}{bold}{\encodingdefault}{\sfdefault}{bx}{n}

\DeclareMathOperator*{\argmax}{arg\,max}

\usepackage{hyperref}
\usepackage{url}
\usepackage{multirow}

\title{An advantage based policy transfer algorithm for reinforcement learning with measures of transferability}

\author{Md Ferdous Alam \email alam.92@osu.edu \\
      \addr The Ohio State University
      \AND
      \name Parinaz Naghizadeh \email parinaz@ucsd.edu \\
      \addr University of California, San Diego 
      \AND
      \name David Hoelzle \email hoelzle.1@osu.edu\\ \addr The Ohio State University
      }

\usepackage{algorithm}
\usepackage{algpseudocode}
\usepackage{graphicx}
\usepackage{amssymb}
\usepackage{enumitem}
\usepackage{amsthm}

\newtheorem{innercustomthm}{Theorem}
\newenvironment{customthm}[1]
  {\renewcommand\theinnercustomthm{#1}\innercustomthm}
  {\endinnercustomthm}
\theoremstyle{definition}
\newtheorem{definition}{Definition}[section]

\begin{document}
\maketitle
\begin{abstract}
Reinforcement learning (RL) enables sequential decision-making in complex and high-dimensional environments through interaction with the environment. In most real-world applications, however, a high number of interactions are infeasible. In these environments, transfer RL algorithms, which can be used for the transfer of knowledge from one or multiple source environments to a target environment, have been shown to increase learning speed and improve initial and asymptotic performance. However, most existing transfer RL algorithms are on-policy and sample inefficient, fail in adversarial target tasks, and often require heuristic choices in algorithm design. This paper proposes an off-policy Advantage-based Policy Transfer algorithm, APT-RL, for fixed domain environments. Its novelty is in using the popular notion of ``advantage'' as a regularizer, to weigh the knowledge that should be transferred from the source, relative to new knowledge learned in the target, removing the need for heuristic choices. Further, we propose a new transfer performance measure to evaluate the performance of our algorithm and unify existing transfer RL frameworks. Finally, we present a scalable, theoretically-backed task similarity measurement algorithm to illustrate the alignments between our proposed transferability measure and similarities between source and target environments. We compare APT-RL with several baselines, including existing transfer-RL algorithms, in three high-dimensional continuous control tasks. Our experiments demonstrate that APT-RL outperforms existing transfer RL algorithms and is at least as good as learning from scratch in adversarial tasks. 
\end{abstract}

\section{Introduction}\label{sec:intro}

A practical approach to implementing reinforcement learning (RL) in sequential decision-making problems is to utilize transfer learning. The transfer problem in reinforcement learning is the transfer of knowledge from one or multiple source environments to a target environment \citep{kaspar2020sim2real, zhao2020sim, bousmalis2018using, peng2018sim, yu2017preparing}. The target environment is the intended environment, defined by the Markov decision process (MDP) $\mathcal{M_T}=\langle \mathcal{S}, \mathcal{A }, \mathcal{R_T}, \mathcal{P_T}\rangle$, where $\mathcal{S}$ is the state-space, $\mathcal{A}$ is the action-space, $\mathcal{R_T}$ is the target reward function and $\mathcal{P_T}$ is the target transition dynamics. The source environment is a simulated or physical environment defined by the tuple $\mathcal{M_S}=\langle \mathcal{S}, \mathcal{A}, \mathcal{R}_\mathcal{S}, \mathcal{P}_\mathcal{S}\rangle$ that, if learned, provides some sort of useful knowledge to be transferred to $\mathcal{M_T}$. While in general $\mathcal{S}$ and $\mathcal{A}$ could be different between the source and target environments, we consider \emph{fixed domain} environments here, where a domain is defined as $\langle\mathcal{S}, \mathcal{A} \rangle$ and is identical in the source and target. 

\begin{figure}[!htb]
\centering
\begin{tabular}{ccc}
    \includegraphics[width=0.15\textwidth]{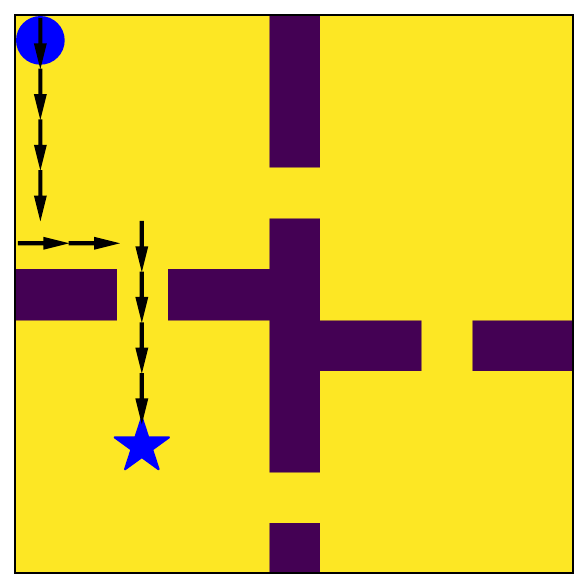} &     
    \includegraphics[width=0.15\textwidth]{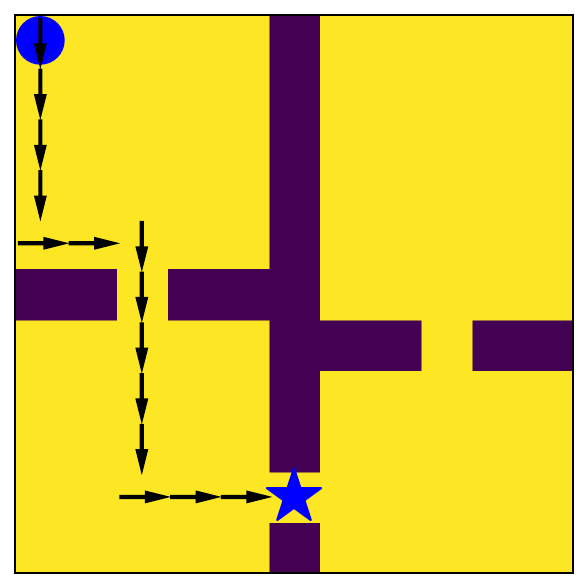} &  
    \includegraphics[width=0.15\textwidth]{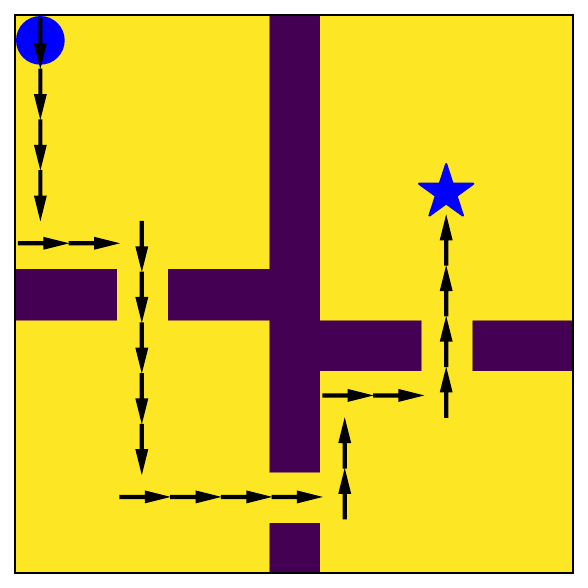}\\     
    (a) Source, $\mathcal{S}$ & (b) Target, $\mathcal{T}_1$ & (c) Target, $\mathcal{T}_2$\\
    \includegraphics[width=0.30\textwidth]{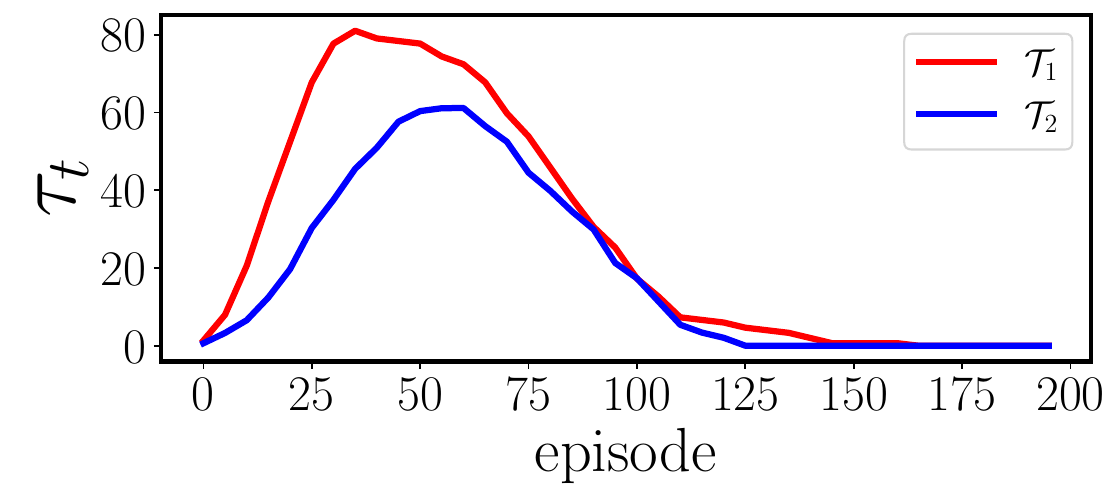} &     
    \includegraphics[width=0.30\textwidth]{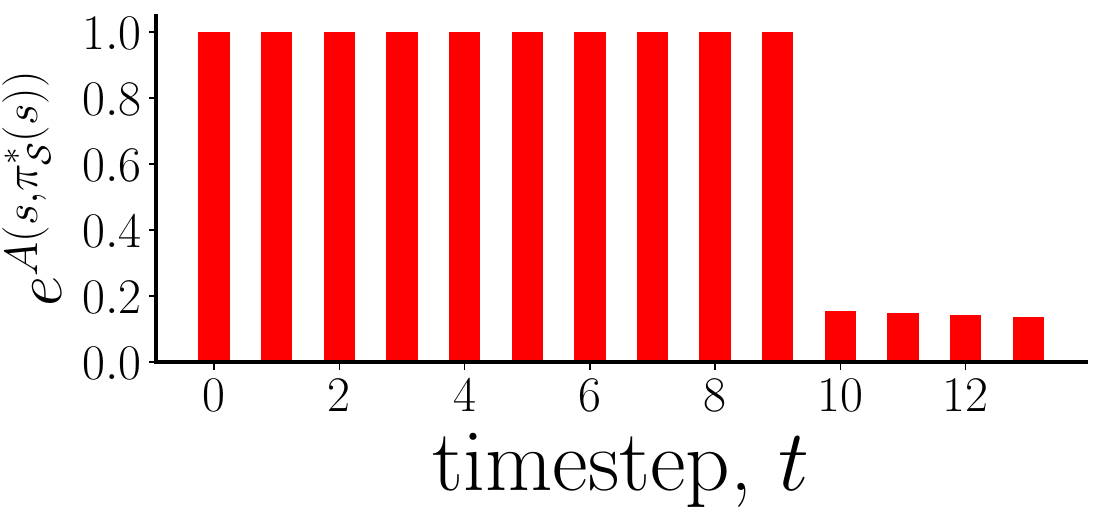} &  
    \includegraphics[width=0.30\textwidth]{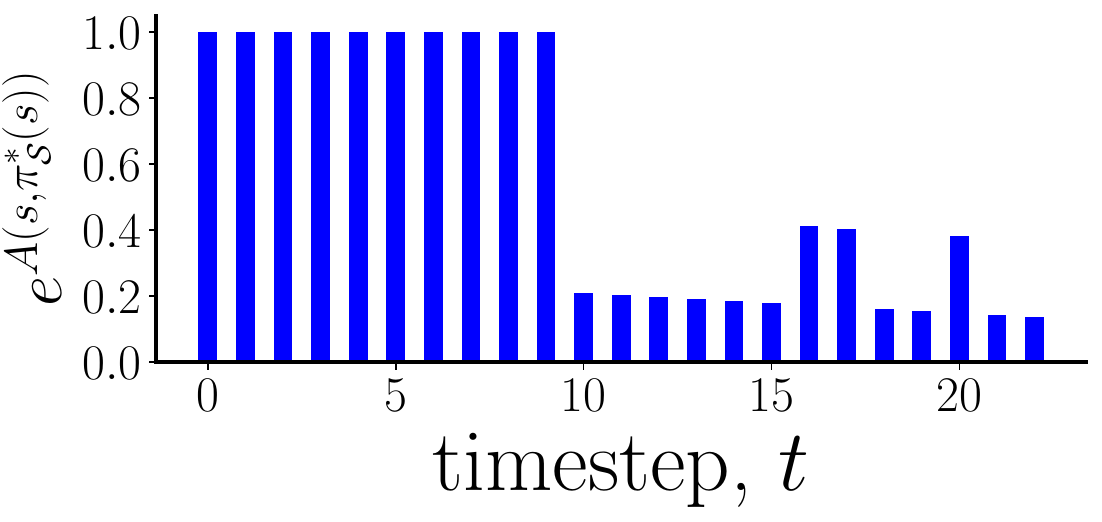}\\ 
    (d) & (e) & (f)
\end{tabular}
\caption{Knowledge transfer in the four-room toy problem: three tasks are presented in (a), (b) and (c), where {\Large\textcolor{blue}{$\bullet$}} represents the starting state and {\Large\textcolor{blue}{$\star$}} represents the goal state; the goal state is moved further in (b) and (c) when compared to (a), and the doorways are also changed slightly. This makes the target task $\mathcal{T}_1$ in (b) more similar to the source task $\mathcal{S}$ in (a) when compared to the other target task $\mathcal{T}_2$ in (c). (d) An evaluation measure, $\tau$, for transfer learning performance in tasks $\mathcal{T}_1$ and $\mathcal{T}_2$ is shown, which calculates performance at each evaluation episode; (e)-(f) influence of source policy on the target tasks $\mathcal{T}_1$ and $\mathcal{T}_2$ are shown in terms of $e^{A(s, a)}$ where $A(s, a) = Q^*_{\mathcal{T}_i}(s, a) - V^*(s)$ is the advantage function in task $\mathcal{T}_i, \text{ and } i = 1, 2$. Note that the action is selected according to the source policy to calculate the advantage, which demonstrates the effect of the source policy on the target.}
\label{fig:toy_problem}
\end{figure}

For effective transfer of knowledge between fixed domain environments, it behooves the user to have $\mathcal{R}_\mathcal{S}$ and $\mathcal{P}_\mathcal{S}$ that are similar to $\mathcal{R_T}$ and $\mathcal{P_T}$, by either intuition, heuristics, or by some codified metric. Consider the single source transfer application to the four-room toy example for two different targets (Fig. \ref{fig:toy_problem}). The objective is to learn a target policy, $\pi^*_{\mathcal{T}}$, in the corresponding target environments, $\mathcal{T}_1$ and $\mathcal{T}_2$ in Fig. \ref{fig:toy_problem}(b) and (c) respectively, by utilizing knowledge from the source task, $\mathcal{S}$ in Fig. \ref{fig:toy_problem}(a), such that the agent applies the optimal or near-optimal sequential actions to reach the goal state. Intuitively, target task $\mathcal{T}_1$ is more similar to the source task $\mathcal{S}$ when compared to target task $\mathcal{T}_2$, and thus we expect that knowledge transferred from $\mathcal{S}$ to $\mathcal{T}_1$ will be comparatively more useful. We propose a formal way to quantify the effectiveness of transfer through a  \emph{transferability evaluation measure}. Our proposed measure, shown as $\tau_1$ for $\mathcal{T}_1$ in Fig. \ref{fig:toy_problem}(d) as a function of evaluation episode $t$, indicates that the source knowledge from $\mathcal{S}$ is highly transferable to the target $\mathcal{T}_1$. In contrast, target task $\mathcal{T}_2$ is less similar to $\mathcal{S}$ and thus knowledge transferred from $\mathcal{S}$ to $\mathcal{T}_2$ may be useful, but should be less effective than that transferred to $\mathcal{T}_1$. Accordingly, the transferability evaluation measure $\tau_2$ for $\mathcal{T}_2$ in Fig. \ref{fig:toy_problem}(d) indicates that the knowledge is useful ($\tau_2 > 0$), but not as useful as the transfer to $\mathcal{T}_1$ ($\tau_1 > \tau_2$). This task similarity measurement approach can provide critical insights about the usefulness of the source knowledge, and as we show, can also be leveraged in comparing the performance of different transfer RL algorithms. We also propose a theoretically-backed, model-based \emph{task similarity measurement} algorithm, and use it to show that our proposed transferability measure closely aligns with the similarities between source and target tasks. Furthermore, we propose a new transfer RL algorithm that uses the popular notion of ``advantage'' as a regularizer, to weigh the knowledge that should be transferred from the source, relative to new knowledge learned in the target.

Our key idea is the following: calculate the advantage function based on actions selected using the source policy, and then use this advantage function as a regularization coefficient to weigh the influence of the source policy on the target. We can observe the effectiveness of this idea in our earlier toy example: Fig. \ref{fig:toy_problem}(e) and (f) show the exponential of the advantage function as the regularization weight. By using the optimal source policy $\pi^*_\mathcal{S}$ to obtain an action, $a$, and calculating the corresponding advantage function $A(s, a)$, it is easy to see that $e^{A(s, a)}$ is lower in $\mathcal{T}_2$ when compared to $\mathcal{T}_1$. This result intuitively means that $\mathcal{S}$ can provide useful guidance for most of the actions selected by the target policy except the last four actions in $\mathcal{T}_1$, whereas, in contrast, the guidance is poor for the last thirteen actions in $\mathcal{T}_2$. We show that this simple yet scalable framework can improve transfer learning capabilities in RL. Our proposed advantage-based policy transfer algorithm is straightforward to implement and, as we show empirically on several continuous control benchmark gym environments, can be at least as good as learning from scratch. 

Our main contributions are the following: 
\begin{itemize}[leftmargin=*]
     \item We propose a novel advantage-based policy transfer algorithm, APT-RL, that enables a data-efficient transfer learning strategy between fixed-domain tasks in RL (Algorithm \ref{alg:apt_rl}). This algorithm incorporates two new ideas: advantage-based regularization of gradient updates, and synchronous updates of the source policy with target data. 
     
     \item We propose a new \emph{relative transfer performance} measure to evaluate and compare the performance of transfer RL algorithms. Our idea extends the previously proposed formal definition of transfer in RL by \citet{lazaric2012transfer}, and unifies previous approaches proposed by \citep{taylor2007transfer, zhu2020transfer}. We provide theoretical support for the effectiveness of this measure (Theorem \ref{thm:relative-transfer}) and demonstrate its use in the evaluation of APT-RL on different benchmarks and against other algorithms (Section \ref{sec:res}). 

     \item We propose a model-based \emph{task similarity measurement} algorithm, and use it to illustrate the relationship between source and target task similarities and our proposed transferability measure (Section \ref{sec:res}). We motivate this algorithm by providing new theoretical bounds on the difference in the optimal policies' action-value functions between the source and target environments in terms of gaps in their environment dynamics and rewards (Theorem \ref{thm:bound}). 
          
     \item We demonstrate the performance of APT-RL in twelve high-dimensional continuous control tasks which are created by varying three OpenAI gym environments, ``halfcheetah'', ``ant'', and ``humanoid''. Across these tasks, we compare APT-RL against four benchmarks: zero-shot source policy transfer, SAC \citep{haarnoja2018soft} without any source knowledge, fine-tuning the source policy with target data, and REPAINT, a state-of-the-art transfer RL algorithm  \citep{tao2020repaint}. Our experiments show that APT-RL outperforms existing transfer RL algorithms on most tasks, and performs as good as learning from scratch in adversarial tasks. 
     
 \end{itemize}
 
The paper is organized as follows: we begin by reviewing related work in \autoref{sec:related}, We propose and explain our off-policy transfer RL algorithm, APT-RL, in \autoref{sec:aptrl}. Next, we propose evaluation measures for APT-RL and a scalable task similarity measurement algorithm between fixed domain environments in \autoref{sec:analysis}. We outline our experiment setup in \autoref{sec:exp}, and use the proposed transferability metric and task similarity measurement algorithm to evaluate the performance of APT-RL and compare it against other algorithms in \autoref{sec:res}. 

\section{Related work}\label{sec:related}
Traditionally, transfer in RL is described as the transfer of knowledge from one or multiple source tasks to a target task to help the agent learn in the target task. Transfer achieves one of the following: a) increases the learning speed in the target task, b) jumpstarts initial performance, and/or c) improves asymptotic performance \citep{taylor2009transfer}. Transfer in RL has been studied extensively in the literature, as evidenced by the three surveys \citep{taylor2009transfer, lazaric2012transfer, zhu2020transfer} on the topic.

The prior literature proposes several approaches for transferring knowledge in RL. One such approach is the transfer of instances or samples \citep{lazaric2008samplestransfer, taylor2008transferring}. Another approach is learning some representation from the source and then transferring it to the target task \citep{taylor2007representation, taylor2005behavior}. Sometimes the transfer is also used in RL for better generalization between several environments instead of focusing on sample efficiency. For example, \citet{barreto2017successor, zhang2017deep} used successor feature representation to decouple the reward and dynamics model for better generalization across several tasks with similar dynamics but different reward functions. Another approach considered policy transfer where the KL divergence between point-wise local target trajectory deviation is minimized and an additional intrinsic reward is calculated at each timestep to adapt to a new similar task \citep{joshi2021adaptive}. In contrast to these works, our proposed approach simply uses the notion of advantage function to transfer \emph{policy parameters} to take knowledge from the source policy to the target task and thus the transfer of knowledge is automated without depending on any heuristic choice.

There have also been different approaches to comparing source and target tasks and evaluating task similarity. A few studies have focused on identifying similarities between MDPs in terms of state similarities or policy similarities \citep{ferns2004metrics, castro2020scalable, agarwal2021contrastive}. A couple of studies also focused on transfer RL where each of the MDP elements is varying \citep{taylor2007cross, gupta2017learning}. Most of these approaches either require a heuristic mapping or consider a high level of similarity between the source and target tasks. In contrast to these works, we develop a scalable task similarity measurement algorithm for fixed domain environments that does not require the learning of the optimal policy. 

A few recent studies have focused on fixed domain transfer RL problems for high-dimensional control tasks \citep{zhang2018decoupling, tao2020repaint}. Most of these studies are built upon on-policy algorithms which require online data collection and tend to be less data efficient than an off-policy algorithm (which we consider here). Although \citet{zhang2018decoupling} discussed off-policy algorithms briefly along with decoupled reward and transition dynamics, a formal framework is absent. Additionally, learning decoupled dynamics and reward models accurately is highly non-trivial and requires a multitude of efforts. More recently, \citet{tao2020repaint} proposed an on-policy actor-critic framework that utilizes the source policy and off-policy instance transfer for learning a target policy. This idea is similar to our approach, but different in two main ways. First, we consider an entirely off-policy algorithm, unlike \citet{tao2020repaint}, and second, our approach does not require a manually tuned hyperparameter for regularization. Additionally, \citet{tao2020repaint} discards samples collected from the target environment that do not follow a certain threshold value which hampers data efficiency. Finally, \citet{tao2020repaint} only considers environments where source and target only vary by rewards and not dynamics. In contrast, we account for varying dynamics, which we believe to be of practical importance for transfer RL applications. 

\section{APT-RL: An off-policy advantage based policy transfer algorithm}\label{sec:aptrl}
In this section, we present our proposed transfer RL algorithm. We explain two main ideas that are novel to this algorithm: advantage-based regularization, and synchronous updates of the source policy. As our proposed algorithm notably utilizes advantage estimates to control the weight of the policy from a source task, we call it \textbf{A}dvantage based \textbf{P}olicy \textbf{T}ransfer in \textbf{RL} (APT-RL). We build upon soft-actor-critic (SAC) \citep{haarnoja2018soft}, a state-of-the-art off-policy RL algorithm for model-free learning. We use off-policy learning to re-use past experiences and make learning sample-efficient. In contrast, an on-policy algorithm would require collecting new samples for each gradient update, which often makes the number of samples required for learning considerably high. Our choice of off-policy learning therefore helps with the scalability of APT-RL to higher dimensional problems, which we will show in the case studies. 

\subsection{Advantage-based policy regularization}
The first new idea in our algorithm is to consider utilizing source knowledge during each gradient step of the policy update. Our intuition is that the current policy, $\pi_\phi$ parameterized by $\phi$, should be close to the source optimal policy, $\pi^*_\mathcal{S}$, when the source can provide useful knowledge. In contrast, when the source knowledge does not aid learning in the target, then less weight should be put on the source knowledge. Based on this intuition, we modify the policy update formula of SAC as follows: we use an additional regularization loss with a temperature parameter, along with the original SAC policy update loss, to control the effect of the added regularization loss. Formally, the policy parameter has the update formula: 
\begin{equation}
     \phi \leftarrow \phi + \alpha_\pi \left[\hat{\nabla}_\phi J_1(\phi) + \beta_t \hat{\nabla}_\phi J_2(\phi) \right]
\label{eq:update}
\end{equation}
\noindent where $J_1(\cdot)$ is the usual SAC policy update loss, which uses soft-Q values instead of Q-values, and Q-function is parameterized by $\theta$ for the dataset $\mathcal{D}_\mathcal{T}$ and learning rate $\alpha_\pi$,
\begin{equation}
    J_1(\phi) =  \mathbb{E}_{\mathbf{s}_t\sim\mathcal{D}_{\mathcal{T}}} \left[ \mathbb{E}_{\mathbf{a}_t \sim \pi_\phi} [\alpha \log(\pi_\phi(\mathbf{a}_t|\mathbf{s}_t)) - Q_\theta(\mathbf{s}_t, \mathbf{a}_t)] \right],
    \label{eq:sac_loss}
\end{equation} 
and $J_2(\cdot)$ is the cross-entropy loss, $\mathcal{H}(\cdot, \cdot)$, between the source policy and the current policy,
\begin{equation}
  J_2(\phi) = \mathcal{H}(\mu_\psi(\mathbf{a}|\mathbf{s}), \pi_\phi(\mathbf{a}|\mathbf{s})) = \mathbb{E}_{\mu_\psi(\mathbf{a}|\mathbf{s})} [-\log \pi_\phi(\mathbf{a}|\mathbf{s})],  \; \; \; 
\end{equation}
\noindent where $\mu_\psi$ represents the optimal source policy $\pi^*_\mathcal{S}$ parameterized by $\psi$. 

\noindent Thus, we are biasing the current policy $\pi_\phi$ to stay close to the source optimal policy $\pi^*_\mathcal{S}$ by minimizing the cross-entropy between these two policies while using the temperature parameter $\beta$ to control the effect of the source policy. Typically, this type of temperature parameter is considered a hyper-parameter and requires (manual) fine-tuning. Finding an appropriate value for this parameter is highly non-trivial and maybe even task-specific. Additionally, if the value of $\beta$ is not appropriately chosen, then the effect of the source policy may be detrimental to learning in the target task. 

To overcome these limitations, we propose an \emph{advantage-based} control of the temperature parameter, $\beta$. The main motivation here is to make the source influence adaptive, which means that we do not treat $\beta$ as a hyper-parameter. The core intuition of our idea is that the second term of \autoref{eq:update} should have more weight when the average action taken according to $\mu_\psi$ is better than the random action. If the source policy provides an action that is worse than a random action in the target, then $\mu_\psi$ should be regularized to have less weight. This is equivalent to taking the difference of the advantages based on the current policy and the source policy, respectively. In addition, we consider the exponential, rather than the absolute value, of this difference, so that the temperature approaches zero when the source provides adversarial knowledge. Formally, our proposed advantage-based temperature parameter is determined as follows:
\begin{equation}
    \begin{split}
    \beta_t  = e^{A^t_\mathcal{S} - A^t_\mathcal{T}}, 
    A^t_\mathcal{T}  = Q_\theta(\mathbf{s}_t, \pi_\phi(\mathbf{s}_t)) - V(\mathbf{s}_t), 
    A^t_\mathcal{S}  = Q_\theta(\mathbf{s}_t, \mu_\psi(\mathbf{s}_t)) - V(\mathbf{s}_t) 
    \end{split}
    \label{eq:adv_old}
\end{equation}
\noindent Note that we can leverage the relationship between the soft-Q values and soft-value functions to represent the advantages from \autoref{eq:adv_old} in a more convenient way that follows from \citep{haarnoja2018soft}:
\begin{equation}
    \begin{split}
    A^t_\mathcal{T} &= Q_\theta(\mathbf{s}_t, \pi_\phi(\mathbf{s}_t)) - \mathbb{E}[Q_\theta(\mathbf{s}_t, \mu_\psi(\mathbf{s}_t)) - \alpha\log \mu_\psi(\mathbf{a}_t|\mathbf{s}_t)]\\
    A^t_\mathcal{S} &= Q_\theta(\mathbf{s}_t, \pi_\phi(\mathbf{s}_t)) - \mathbb{E}[Q_\theta(\mathbf{s}_t, \pi_\phi(\mathbf{s}_t)) - \alpha\log \pi_\phi(\mathbf{a}_t|\mathbf{s}_t)]
    \end{split}    
    \label{eq:adv}
\end{equation}
    
\subsection{Synchronous update of the source policy}

We propose an additional improvement over the advantage-based policy transfer idea to further improve sample efficiency. As we consider a parameterized source optimal policy, $\mu_\psi$, it is possible to update the parameters of the source policy with the target data, $\mathcal{D}_\mathcal{T}$, by minimizing the SAC loss. The benefits of this approach is two-fold: 1) If the source optimal policy provides useful information to the target, then this will accelerate the policy optimization procedure by working as a regularization term in \autoref{eq:update}, and 2) This approach enables {sample transfer} to the target policy. This is because the initial source policy is learned using the source data; when this source policy is further updated with the target data, it can be viewed as augmenting the source dataset with the latest target data. Formally, the source policy will be updated as follows: $\psi \leftarrow \psi + \alpha_\psi \hat{\nabla}_\psi J_1(\psi)$, where $J_1(\psi)$ is the typical SAC loss for the source policy with parameters $\psi$ and learning rate $\alpha_\psi$. The pseudo-code for APT-RL is shown in Algorithm \ref{alg:apt_rl}. 

\begin{algorithm}
    \caption{APT-RL: \textbf{A}dvantage based \textbf{P}olicy \textbf{T}ransfer in \textbf{R}einforcement \textbf{L}earning}
    \label{alg:apt_rl}
    
\begin{algorithmic}[1]
    \State \textbf{Given:} parameterized source optimal policy, $\mu_\psi$, source learning step $\alpha_\mathcal{S}$
    \State \textbf{Initialize:} current target policy, $\pi_\phi$, target buffer $\mathcal{D}_\mathcal{T} = \emptyset$
    \For{each iteration}
        \For{{{each target environment step}}}
            \State $\mathbf{a} \sim \pi_\phi(\mathbf{a}_t|\mathbf{s}_t)$
            \State $\mathbf{s}' \sim p(\mathbf{s}'|\mathbf{s}, \mathbf{a})$
            \State $\mathcal{D}_\mathcal{T} \leftarrow \mathcal{D}_\mathcal{T} \cup \{(\mathbf{s}, \mathbf{a}, \mathcal{R}(\mathbf{s}, \mathbf{a}), \mathbf{s}')\}$
        \EndFor
        
        \For{$G$ gradient updates}
            \State $\psi \leftarrow \psi + \alpha_\mathcal{S} \hat{\nabla}_\psi J_1(\psi)$ \text{ where $J_1$ is defined in \autoref{eq:sac_loss}}        
            \State calculate $A_\mathcal{T}^t$ and $A^t_\mathcal{S}$ using \autoref{eq:adv}
            \State $\beta_t \leftarrow e^{A^t_\mathcal{S} - A^t_\mathcal{T}}$
            \State  $\phi \leftarrow \phi + \alpha_\pi \left[ \hat{\nabla}_\phi J_1(\phi) + \beta_t \hat{\nabla}_\phi J_2(\phi) \right]$ 
        \EndFor
    \EndFor 
\end{algorithmic}
\end{algorithm}

\section{An evaluation framework for transfer RL}\label{sec:analysis}

To formally quantify the performance of APT-RL, including against other algorithms, in this section, we propose notions of transferability and task similarity, as discussed in Section \ref{sec:intro}. 
First, we propose a formal notion of transferability and use this notion to calculate a ``relative transfer performance'' measure. We demonstrate how this measure can be utilized to assess and compare the performance of APT-RL and similar algorithms. Then, we propose a scalable task similarity measurement algorithm for high-dimensional environments. We motivate this algorithm by providing a new theoretical bound on the difference in the optimal policies action-value functions between the source and target environments in terms of gaps in their environment dynamics and rewards. This task similarity measurement algorithm can be used to identify the best source task for transfer. Further, we use this algorithm in the experiment section, to illustrate how our proposed relative transfer performance closely aligns with the similarities between the source and target environment.

\subsection{A measure of transferability}
We formally define \emph{transferability} as a mapping from stationary source knowledge and non-stationary target knowledge accumulated until timestep $t$, to learning performance, $\rho_t$. 

\begin{definition}[Single-task transferability]
Let $\mathcal{K}_\mathcal{S}$ be the transferred knowledge from a source task $\mathcal{M}_\mathcal{S}$ to a target task $\mathcal{M}_\mathcal{T}$, and let $\mathcal{K}_{\mathcal{T}, t}$ be the available knowledge in $\mathcal{M}_\mathcal{T}$ at timestep $t$. Let $\rho_t$ denote a measure that evaluates the learning performance in $\mathcal{M}_\mathcal{T}$ at timestep $t$. Then, transferability is defined as the mapping
$$\Lambda: \mathcal{K}_\mathcal{S} \times \mathcal{K}_{\mathcal{T}, t} \rightarrow \rho_t~.$$
\end{definition}

Intuitively, this means that $\Lambda(\cdot)$ takes prior source knowledge and accumulated target knowledge to evaluate the learning performance in the target task. 

As an example, if the collection of source data samples $\mathcal{D}_\mathcal{S}$ are utilized as the transferred knowledge and the average returns using the latest target policy, $G_t = \mathbb{E}^{\pi_{\mathcal{T}, t}}\left[\sum_{k=0}^{H}r_k\right]$, is used as the evaluation performance of a certain transfer algorithm $i$, then the transferability of algorithm $i$ at episode $t$, can be written as the following, $\Lambda_i: \mathcal{D}_\mathcal{S} \times \mathcal{D}_{\mathcal{T}, t} \rightarrow G_t$.

\begin{table}
\centering
\begin{tabular}{ccc}
\hline
    source knowledge, $\mathcal{K}_\mathcal{S}$ & target knowledge, $\mathcal{K}_{\mathcal{T}, t}$ & evaluation measure, $\rho_t$ \\ 
    \hline     
    samples, $\mathcal{D}_\mathcal{S}$ & samples, $\mathcal{D}_{\mathcal{T}, t}$ & average returns, $G_t = \mathbb{E}^{\pi_{\mathcal{T}, t}}\left[\sum_{k=0}^H r_k\right]$ \\ 
    
    policy, $\mu_\psi$ & current policy, $\pi_{\mathcal{T}, t}$ & samples required to obtain reward threshold, $n_\text{th}$ \\ 
    
    models $\mathcal{R}_\mathcal{S}, \mathcal{P}_\mathcal{S}$ & models, $\mathcal{R}_{\mathcal{T}, t}, \mathcal{P}_{\mathcal{T}, t}$ & area under the reward curve, $\Delta_{t}$\\ 
    
    value functions $Q^*_\mathcal{S}, V^*_\mathcal{S}$ & value functions $Q_{\mathcal{T}, t}, V_{\mathcal{T}, t}$ & samples required to obtain asymptotic returns, $n_\infty$\\
    \hline 
\end{tabular}
\caption{A list of potential source knowledge, target knowledge and evaluation performance. This list unifies previous approaches proposed by \citep{taylor2007transfer, zhu2020transfer}. Note that target knowledge and evaluation performance is represented for the $t^{th}$ episode, and that $\pi_{\mathcal{T}, t}$ denotes the optimal target policy at the evaluation episode $t$.}
\label{table:transferability}
\end{table}

\noindent Notice that we leave the choice of input and output of this mapping as user-defined task-specific parameters. Any traditional transfer methods can be represented using the idea of transferability. Potential choices for source and target knowledge and evaluation measures are listed in Table \ref{table:transferability}. Our idea extends the previously proposed formal definition of transfer in RL by \citet{lazaric2012transfer}, and unifies previous approaches proposed by \citet{taylor2007transfer, zhu2020transfer}. Expressing transfer learning algorithms in terms of this notion of transferability has a number of advantages. First, this problem formulation can be easily extended to unify other important RL settings. For example, this definition can be extended to offline RL \citep{levine2020offline} by considering $\mathcal{K}_\mathcal{S} = \mathcal{D}_{\text{source}},  \text{ and } \mathcal{K}_{\mathcal{T}, t} = \emptyset, \ \forall t.$ Second, the comparison of two transfer methods becomes convenient if they have the same evaluation criteria. For instance, one way to construct evaluation criteria may be to use sample complexity in the target task to achieve a desired return. Subsequently, the transferability measure can be used to assess ``relative transfer performance'', which can act as a tool for comparing two different transfer methods conveniently.  

\begin{definition}[Relative transfer performance, $\tau$]
Given the transferability mapping of algorithm $i$, $\Lambda_i$, the relative transfer performance is defined as the difference between the corresponding learning performance $\rho_t^i$ and learning performance from a base RL algorithm $\rho_t^b$ at evaluation episode $t$. Formally, $\tau_{t} = \rho^i_{t} - \rho^b_{t}~$, where the base RL algorithm represents learning from scratch in the target task (meaning $\mathcal{K}_\mathcal{S}=\emptyset$), and $\rho_t^i$, $\rho_t^b$ are the same evaluation criteria of learning performances. 
\end{definition}  

\subsubsection{Theoretical support} 
We first formally show that, with an appropriate definition of the evaluation measure, non-negative relative transfer performance leads to a policy in the target task which is at least as good as learning from scratch.

\begin{customthm}{1}\label{thm:relative-transfer}{\textbf{(Relative transfer performance and policy improvement)}} Consider $\rho_t^i = \mathbb{E}^{\pi_{i, t}}\left[\sum_{k=0}^H r_k|\mathbf{s}_0\right]$ for policy $\pi_i$ and $\rho_t^b = \mathbb{E}^{\pi_{b, t}}\left[\sum_{k=0}^H r_k|\mathbf{s}_0\right]$ for policy $\pi_b$, where $\mathbf{s}_0$ is the starting state and each policy is executed for $H$ timesteps. Then, the learned policy, $\pi_{i, t}$ using algorithm $i$, in the target at episode $t$ is at least as good as the source optimal policy $\pi_{b, t}$ if $\tau_t \geq 0$.  
\end{customthm}
The proof can be found in the Appendix \ref{app:thm1}. 

\subsubsection{Revisiting the toy problem}
We also leverage the toy example presented in Fig. \ref{fig:toy_problem} to explain the proposed concepts. The performance evaluation is chosen as the average returns, collected from an evaluation episode $t$, that is $\rho_t = \sum_{k=0}^H r_k$. For transfer, we choose the low-level direct knowledge Q-values. At first, we initialize the target Q-values with pre-trained source Q-values, $Q^*_\mathcal{S}$. Thus, at each episode, $t$, the updated Q-values in the target are a combination of both source and target knowledge. Thus, the transferability mapping can be expressed as $\Lambda_\text{Q-learning}: Q^*_\mathcal{S} \times Q_{\mathcal{T}, t} \rightarrow \mathbb{E}^{\pi_{\mathcal{T}, t}}\left[\sum_{k=0}^H r_k\right]$. We calculate $\rho_t$ after every $10$ timestep by executing the greedy policy from the updated Q-values for a fixed time horizon. Relative transfer performance, $\tau_t$, remains non-negative for both the target tasks $\mathcal{T}_1$ and $\mathcal{T}_2$, for up to around $125$ evaluation episodes. Intuitively this means that the learning performance of both policies is better than a base algorithm for all of the evaluation episodes. Also, $\tau_t$ is higher for $\mathcal{T}_1$ than $\mathcal{T}_2$ which means that transferring knowledge from $\mathcal{S}$ leads to better learning performance in $\mathcal{T}_1$ compared to $\mathcal{T}_2$. This can be explained by the fact that the dynamics in $\mathcal{T}_1$ are more similar to $\mathcal{S}$ than $\mathcal{T}_2$.

\subsection{Measuring task similarity}
As seen in the toy problem above, a measure of task similarity would help us illustrate the close alignment of our proposed transferability metric with the similarities of the source and target tasks. Beyond this, measuring task similarity can provide additional insights into why a particular source task is more appropriate to transfer knowledge to the target task. Motivated by these, we propose an algorithm for measuring task similarity in this section. 

\subsubsection{Theoretical motivation}
To motivate the idea behind our proposed algorithm, we first investigate theoretical bounds on the expected discrepancies between the policies learned in the source and target environments. One effective way for this analysis is to calculate the upper bound on differences between the optimal policies' action-values. Previously, action-value bounds have been proposed for similar problems by \cite{csaji2008value, abdolshah2021new}. We extend these ideas to the transfer learning setting where we derive the bound between the target action-values under target optimal policy, $\pi^*_\mathcal{T}$ and target action-values under source optimal policy, $\pi^*_\mathcal{S}$. 

\begin{customthm}{2}\label{thm:bound}{\textbf{(Action-value bound between fixed-domain environments)}} If $\pi^*_\mathcal{S}$ and $\mathcal{\pi^*_\mathcal{T}}$ are the optimal policies in the MDPs $\mathcal{M}_\mathcal{S} = \langle \mathcal{X}, \mathcal{A}, \mathcal{R}_\mathcal{S}, \mathcal{P}_\mathcal{S}\rangle$ and $\mathcal{M}_\mathcal{T} = \langle \mathcal{X}, \mathcal{A}, \mathcal{R}_\mathcal{T}, \mathcal{P}_\mathcal{T}\rangle$ respectively, then the corresponding action-value functions is upper bounded by 
\begin{equation}
    ||\mathbf{Q}^{\pi^*_\mathcal{T}}_\mathcal{T} -  \mathbf{Q}^{\pi^*_\mathcal{S}}_\mathcal{T}||_\infty \leq \frac{2\delta^r_{\mathcal{S}\mathcal{T}}}{1-\gamma} + \frac{2\gamma \delta^{TV}_{\mathcal{S}\mathcal{T}}(R_{max, \mathcal{S}} + R_{max, \mathcal{T}})}{(1-\gamma)^2}
    \label{eq:bound-new}
\end{equation}

where $\delta^r_{\mathcal{S}\mathcal{T}} = ||\mathcal{R}_\mathcal{S}(\mathbf{s}, \mathbf{a})- \mathcal{R}_\mathcal{T}(\mathbf{s}, \mathbf{a}))||_\infty$, $\delta^{TV}_{\mathcal{S} \mathcal{T}}$ is the total variation distance between $\mathcal{P}_\mathcal{S}$ and $\mathcal{P}_\mathcal{T}$, $\gamma$ is the discount factor and $R_{max, \mathcal{S}} = ||\mathcal{R}_\mathcal{S}(\mathbf{s}, \mathbf{a})||_\infty$, $R_{max, \mathcal{T}} = ||\mathcal{R}_\mathcal{T}(\mathbf{s}, \mathbf{a})||_\infty$.
\end{customthm}

The proof of this theorem can be found in the Appendix \ref{app:thm2}. Note that as we propose APT-RL for fixed-domain environments, the bound can be expressed in terms of differences in the remaining environment parameters: the total variation distance between the source and target transition probabilities, and the maximum reward difference between the source and the target. Intuitively this bound means that a lower total variation distance between the transition dynamics can provide a tighter bound on the deviation between the action-values from the target and source optimal policies. Similarly, having a smaller reward difference also helps in getting lower action-value deviations in the target. Also note that, if the reward function or transition dynamics remain identical between source and target, then the corresponding term on the right side of \autoref{eq:bound-new} vanishes. 

Next, motivated by this bound, we propose a task similarity measurement algorithm that assesses the differences between source and target dynamics and rewards in order to evaluate their similarity. 

\subsubsection{A model-based task similarity measurement algorithm} \label{sec:task_sim}
Previous attempts for measuring task similarity include behavioral similarities in MDP in terms of state-similarity or bisimulation metric, and policy similarity \citep{ferns2004metrics, castro2020scalable, agarwal2021contrastive}. Calculating such metrics in practice is often challenging due to scalability issues and computation limits. Additionally, our key motivation is to find a similarity measurement that does not require solving for the optimal policy \textit{apriori}, as the latter is often the key challenge in RL. To this end, we propose a new model-based method for calculating similarities between tasks. 

We propose an encoder-decoder based deep neural network model at the core of this idea. For any source or target task, a dataset, $\mathcal{D} = \{(\mathbf{s}, \mathbf{a}, r, \mathbf{s}')\}$, is collected by executing a random policy. Next, a dynamics model, $f_\mathcal{P}(\mathbf{s}, \mathbf{a})$, is trained by minimizing the mean-squared-error (MSE) loss using stochastic gradient descent, $\mathcal{L}_\text{dyn} = ||\mathbf{s}' - f_\mathcal{P}(\mathbf{s}, \mathbf{a})||_2$. Similarly, a reward model, $f_\mathcal{R}(\mathbf{s}, \mathbf{a})$ is trained using the collected data to minimize the following MSE loss, $\mathcal{L}_\text{rew} = ||r - f_\mathcal{R}(\mathbf{s}, \mathbf{a})||_2$. The encoder portion of the neural network model encodes state and action inputs into a latent vector. Then, the decoder portion uses this latent vector for the prediction of the next state or reward. We consider decoupled models for this purpose, meaning that we learn separate models for reward and transition dynamics from the same dataset. This allows us to identify whether only reward or transition dynamics or both vary between tasks. Once these models are learned, the source model is used to predict the target data and calculate the $L_2$ distance between the predicted and actual data as the similarity error. If $\xi_k^\mathcal{P}$ and $\xi_k^\mathcal{R}$ are the similarity errors in target dynamics and rewards,  respectively, we can calculate the dynamics and reward similarity separately as follows, 
\begin{equation}
\begin{aligned}    
    \text{dynamics similarity:  } \Xi^\mathcal{P}_{\mathcal{S},\mathcal{T}} = \frac{1}{|\mathcal{D}_\mathcal{T}|}\sum_{k=1}^{|\mathcal{D}_\mathcal{T}|} \xi_k^\mathcal{P}, 
    \text{reward similarity: } \Xi^\mathcal{R}_{\mathcal{S}, \mathcal{T}} = \frac{1}{|\mathcal{D}_\mathcal{T}|}\sum_{k=1}^{|\mathcal{D}_\mathcal{T}|} \xi_k^\mathcal{R}
\end{aligned}
\end{equation}
Our approach is summarized in Algorithm \ref{alg:task_sim}. This approach can be viewed as a modern version of \citep{ammar2014automated}, but instead of using Restricted Boltzmann Machines, we use deep neural network-based encoder-decoder architecture to learn the models, and we do it in a decoupled way. 

\begin{algorithm}.
    \caption{Model-based task similarity measurement}
    \label{alg:task_sim}
\begin{algorithmic}[1]
    \State Collect $m$ data samples from source $\mathcal{M}_\mathcal{S}$ using a random policy, $\mathcal{D}_\mathcal{S} = \{(\mathbf{s}_\mathcal{S}, \mathbf{a}_\mathcal{S},  r_\mathcal{S}, \mathbf{s}_\mathcal{S}')\}$ 
    \State Collect $m$ data samples from target $\mathcal{M}_\mathcal{T}$ using a random policy, $\mathcal{D}_\mathcal{T} = \{(\mathbf{s}_\mathcal{T}, \mathbf{a}_\mathcal{T},  r_\mathcal{T}, \mathbf{s}_\mathcal{T}')\}$ 
    \State Learn dynamics models, $f_\mathcal{P}^\mathcal{S}(\mathbf{s}, \mathbf{a}), f_\mathcal{P}^\mathcal{T}(\mathbf{s}, \mathbf{a})$ using $\mathcal{D}_\mathcal{S}$ and $\mathcal{D}_\mathcal{T}$ and minimizing $\mathcal{L}_\text{dyn} = ||\mathbf{s}' - f_\mathcal{P}(\mathbf{s}, \mathbf{a})||_2$
    \State Learn  reward models, $f_{\mathcal{R}_\mathcal{S}}(\mathbf{s}, \mathbf{a}), f_{\mathcal{R}_\mathcal{T}}(\mathbf{s}, \mathbf{a})$ using $\mathcal{D}_\mathcal{S}$ and $\mathcal{D}_\mathcal{T}$ and minimizing $\mathcal{L}_\text{rew} = ||r - f_\mathcal{R}(\mathbf{s}, \mathbf{a})||_2$  
    \For {each $(\mathbf{s}_\mathcal{T}, \mathbf{a}_\mathcal{T},  r_\mathcal{T}, \mathbf{s}_\mathcal{T}') \in \mathcal{D}_\mathcal{T}$}
        \State $\hat{\mathbf{s}}_\mathcal{T}' = f_{\mathcal{P}_\mathcal{T}}(\mathbf{s}_\mathcal{T}, \mathbf{a}_\mathcal{T}), \hat{r}_\mathcal{T} = f_{\mathcal{R}_\mathcal{T}}(\mathbf{s}_\mathcal{T}, \mathbf{a}_\mathcal{T})$
        \State $\xi_k^\mathcal{P} = ||\hat{\mathbf{s}}_\mathcal{T}' - \mathbf{s}_\mathcal{T}'||_2, \xi_k^\mathcal{R} = ||\hat{r}_\mathcal{T} - r_\mathcal{T}||_2$
    \EndFor
    \State dynamics similarity, $\Xi^\mathcal{P}_{\mathcal{S},\mathcal{T}} = \frac{1}{|\mathcal{D}^\mathcal{T}|}\sum_{k=1}^{|\mathcal{D}^\mathcal{T}|} \xi_k^\mathcal{P}$
    \State reward similarity, $\Xi^\mathcal{R}_{\mathcal{S}, \mathcal{T}} = \frac{1}{|\mathcal{D}^\mathcal{T}|}\sum_{k=1}^{|\mathcal{D}^\mathcal{T}|} \xi_k^\mathcal{R}$
\end{algorithmic}
\end{algorithm}
\section{Experiment Setup}\label{sec:exp}
We apply the described methods to three popular high-dimensional continuous control benchmark gym environments \citep{brockman2016openai}: 1) `HalfCheetah-v3', 2) `Ant-v3', and 3) `Humanoid-v3'. The vanilla Gym environments are not suitable for transfer learning settings. Thus we create four different perturbations of the dynamics of each environment. The source HalfCheetah-v3 environment has the standard gym values for damping and the four target environments have different values
of damping increased gradually in each task. The least similar task has the highest damping values in the joints. For the Ant-v3 environment, the source task has the standard gym robot and the four target environments have varied dynamics by changing the leg lengths of the robot. Similarly, for the Humanoid-v3 environment, the source task has the standard gym robot and the four target environments have varied dynamics by changing the leg length as well as the size of the hand and legs. For the HalfCheetah-v3 environment, we create additional four tasks by changing the reward function to show reward variation. Details of the environments and the algorithm hyperparameters can be found in Appendices \ref{app:exp_details} and \ref{app:params}.

For each environment, the knowledge transferred is the source optimal policy, data collected from the target task is used as the target knowledge, and average returns during each evaluation episode are used as the performance evaluation metric, $\Lambda_i: \pi^*_\mathcal{S} \times \mathcal{D}_{\mathcal{T}, t} \rightarrow \mathbb{E}^{\pi_{\mathcal{T}, t}}\left[\sum_k r_k\right]$. To obtain the source optimal policy, the SAC algorithm is utilized to train a policy from scratch in each source environment. In each of these experiments, we perform and compare our algorithm APT-RL against one of the recent benchmarks on transfer RL, the REPAINT algorithm proposed by \citet{tao2020repaint}. We also compare APT-RL against zero-shot source policy transfer, policy fine-tuning on the target task, and SAC without any source knowledge. 

\section{Experiment Results}\label{sec:res}
\subsection{Task similarity}
\begin{figure}[!htb]
\centering
\begin{tabular}{cccc}
\includegraphics[width=0.20\textwidth]{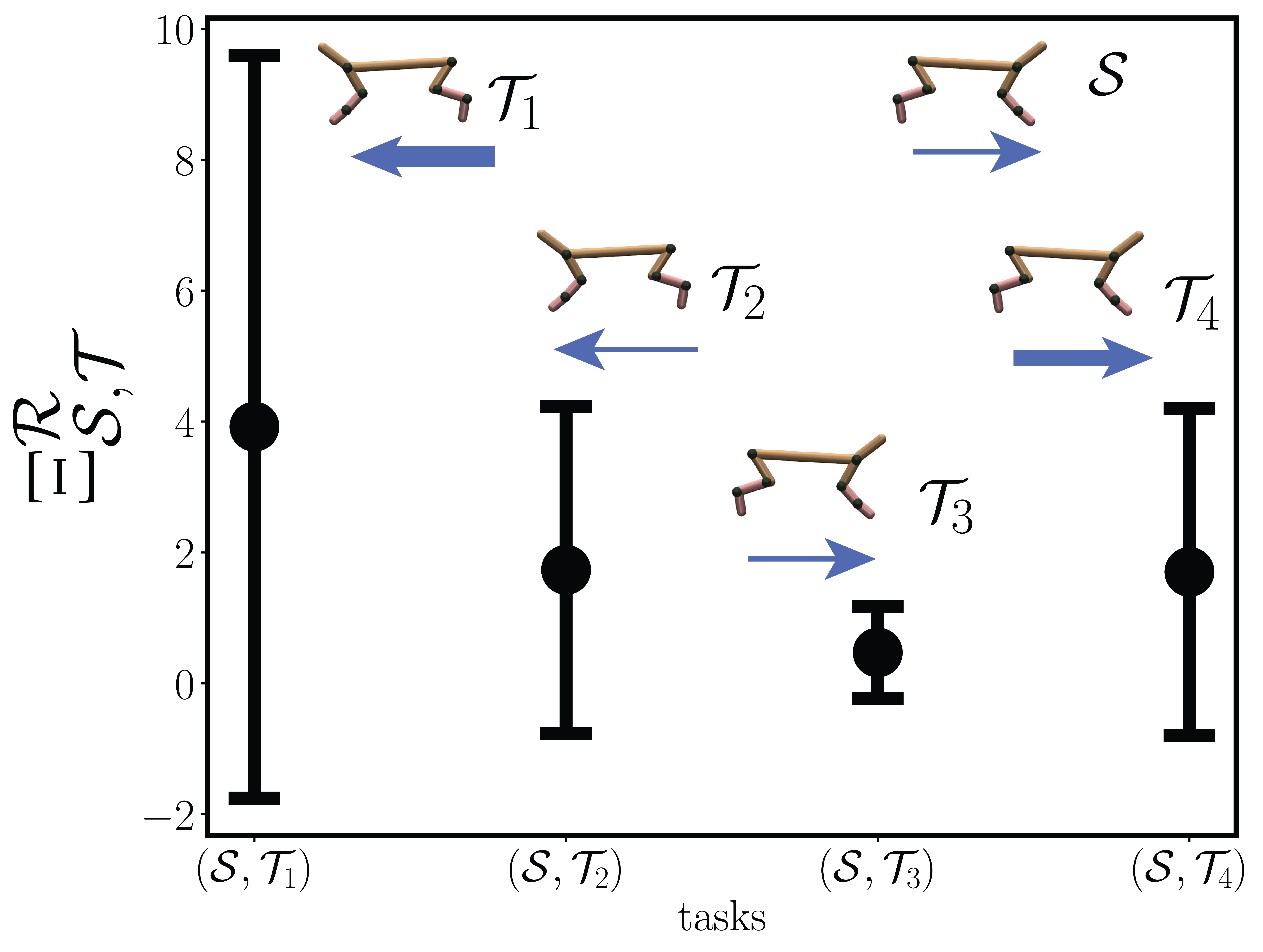} &     \includegraphics[width=0.20\textwidth]{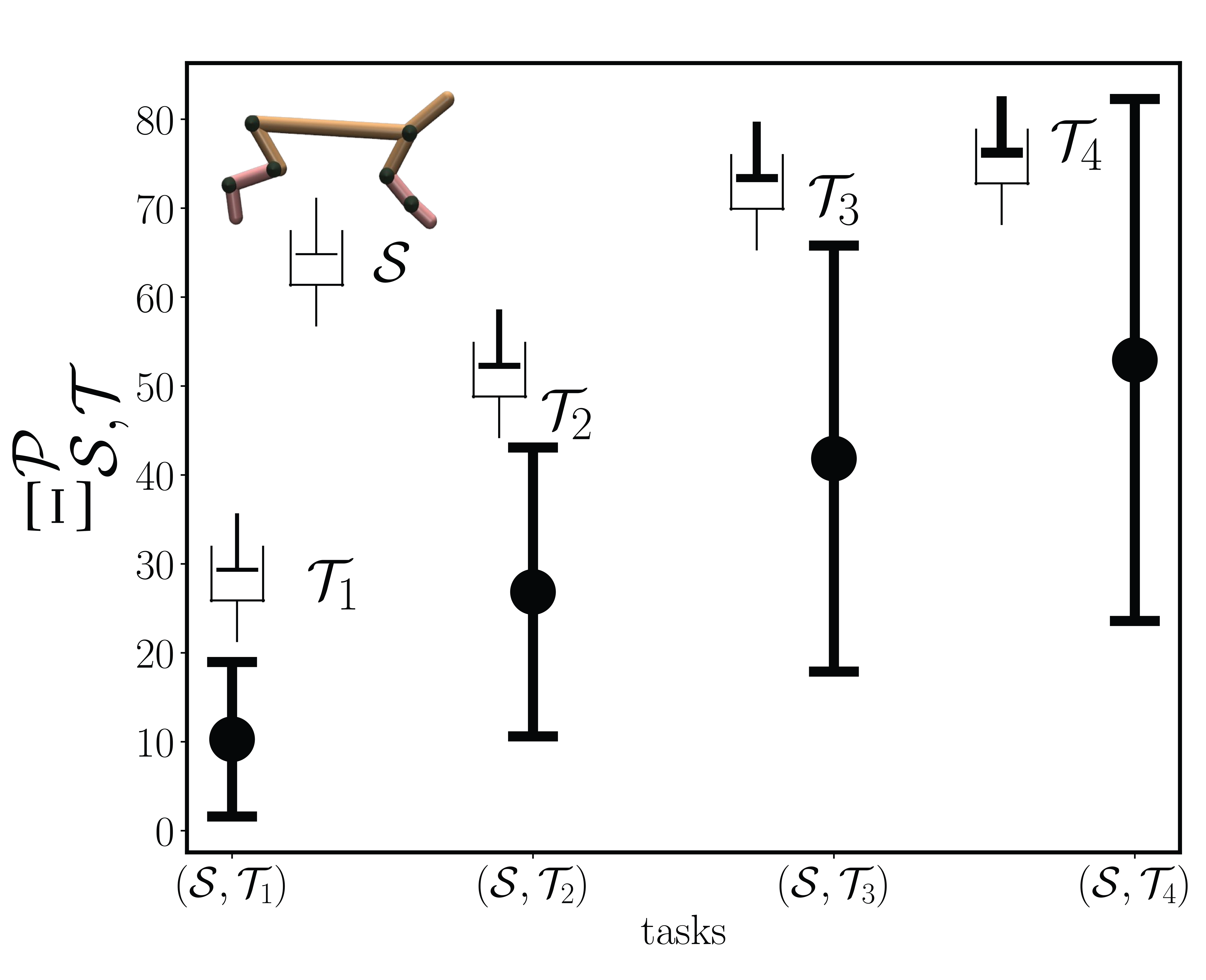} &   
\includegraphics[width=0.20\textwidth]{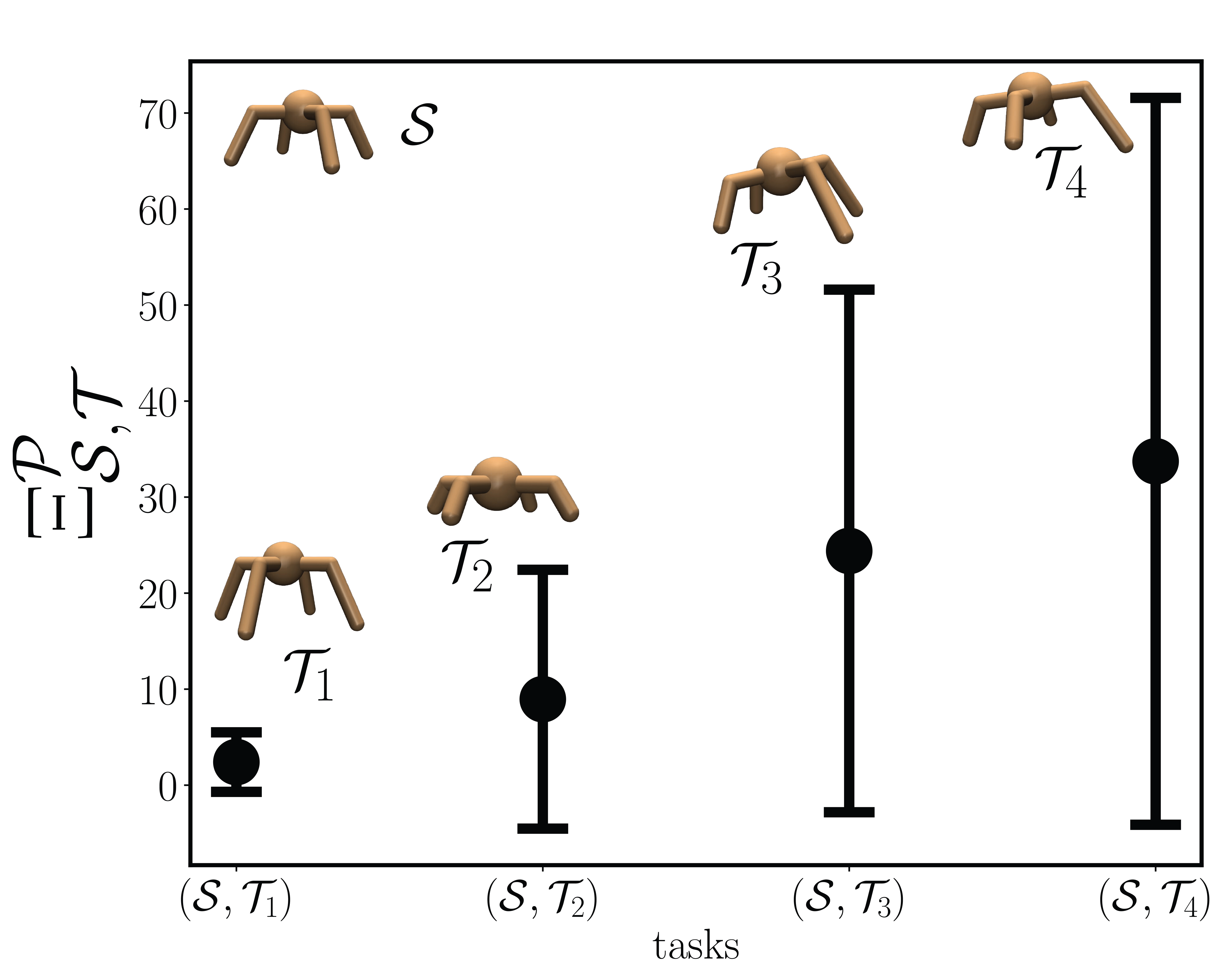} &
\includegraphics[width=0.210\textwidth]{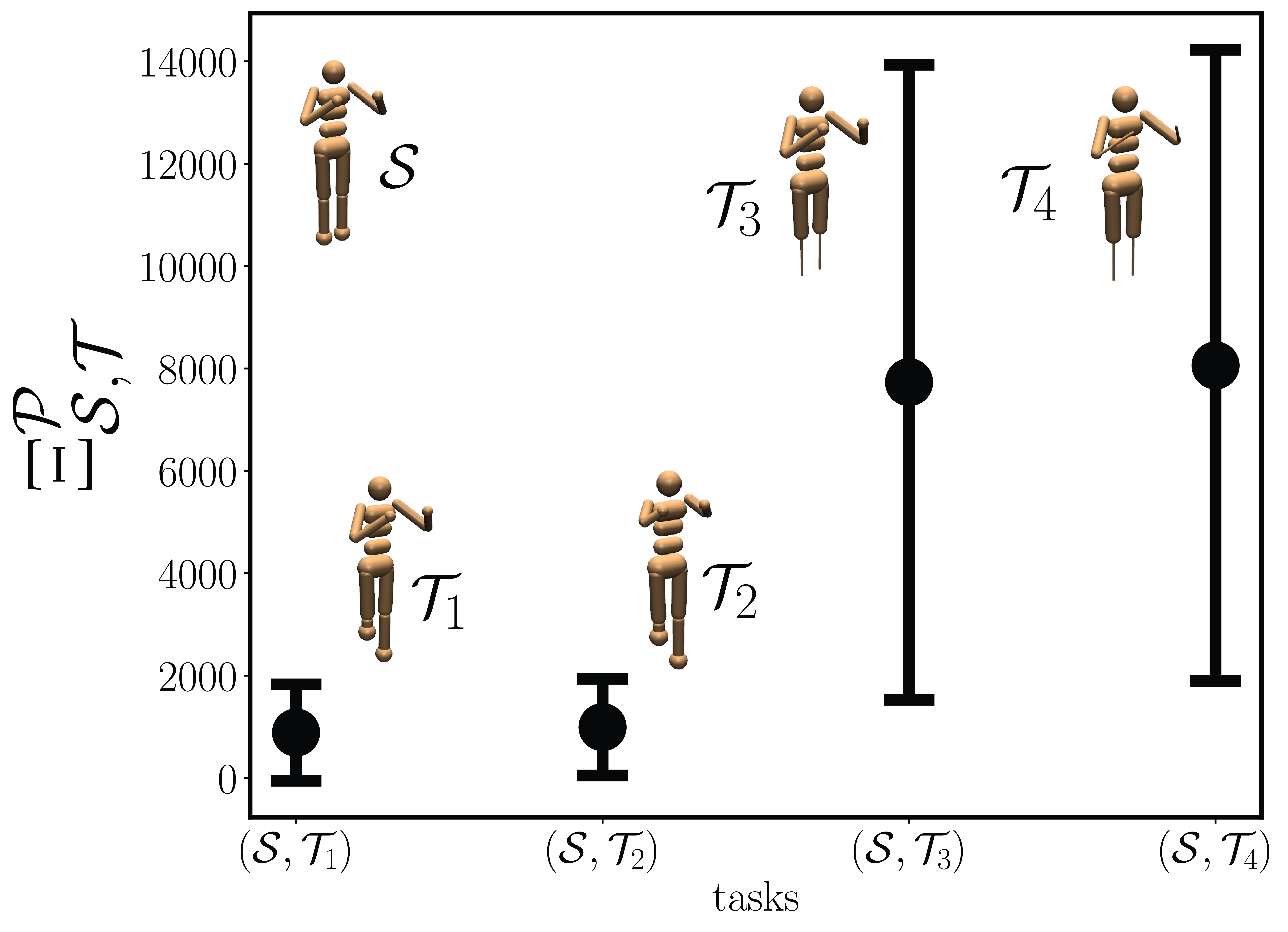} \\ 
    (a) Halfcheetah (reward) & (b) Halfcheetah (dynamics) & (c) Ant (dynamics) & (d) Humanoid (dynamics)
\end{tabular}
\caption{\textbf{Task dissimilarity:} Empirical task similarity between several variations of Half-cheetah, Ant, and Humanoid environments}.
\label{fig:task_sim}
\end{figure}

We leverage Algorithm \ref{alg:task_sim} to calculate the task similarity between the source and each of the target tasks. The empirical task similarity is shown in Fig. \ref{fig:task_sim}. Fig. \ref{fig:task_sim}(a) shows the similarity in tasks for the half-cheetah environment with varying rewards. For reward similarity, we can see the highest dissimilarity when the robot is provided a negative reward instead of a positive reward. Similarly, for the half-cheetah environment with varying dynamics in Fig. \ref{alg:task_sim}(b), we can see an approximately linear trend in the similarity of the dynamics. This makes sense as the change in joint damping is gradual and constant. In Fig. \ref{fig:task_sim}(c) we show the task similarity in the Ant environment with varying dynamics. As we change the dynamics of each of the target environments by changing the length of the legs of the robot, the similarity between each target and the source task reduces monotonically. Finally, task similarity of the humanoid environments are shown in Fig. \ref{fig:task_sim}(d) where first two tasks are relatively more similar to the source and the final two tasks are less similar. 

\subsection{Transferability of APT-RL, $\Lambda_\text{APT-RL}$}
\subsubsection{Half-cheetah-v3}
\begin{figure}[!htb]
\centering
\begin{tabular}{@{}c@{}c@{}c@{}c@{}c@{}}
    \rotatebox{90}{\parbox{3cm}{\centering Half-Cheetah}} & 
    \includegraphics[width=0.25\textwidth]{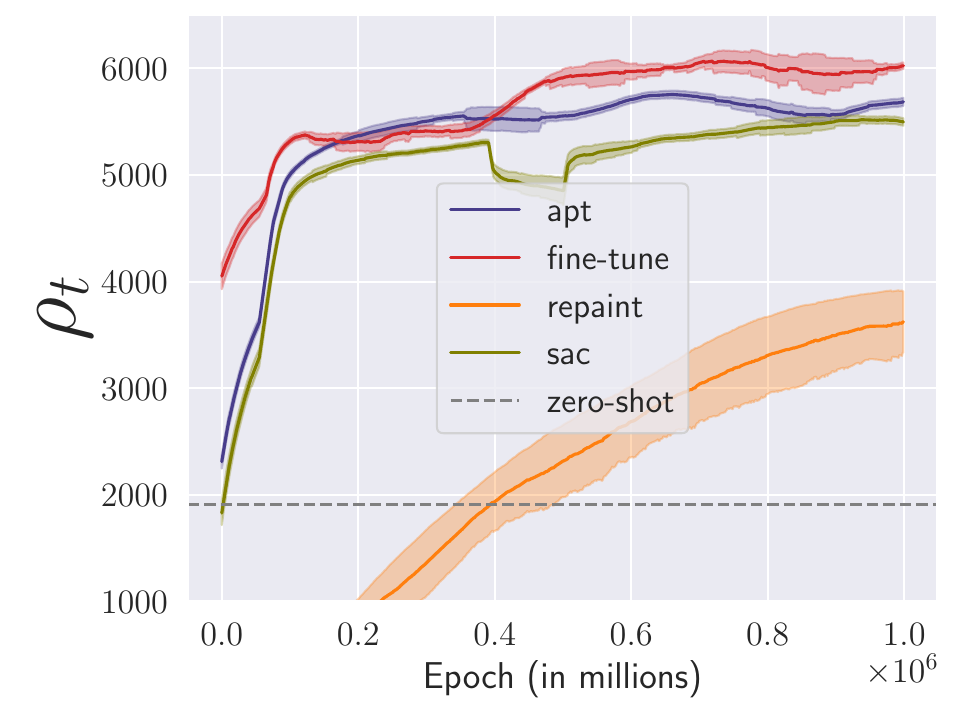} &  
    \includegraphics[width=0.25\textwidth]{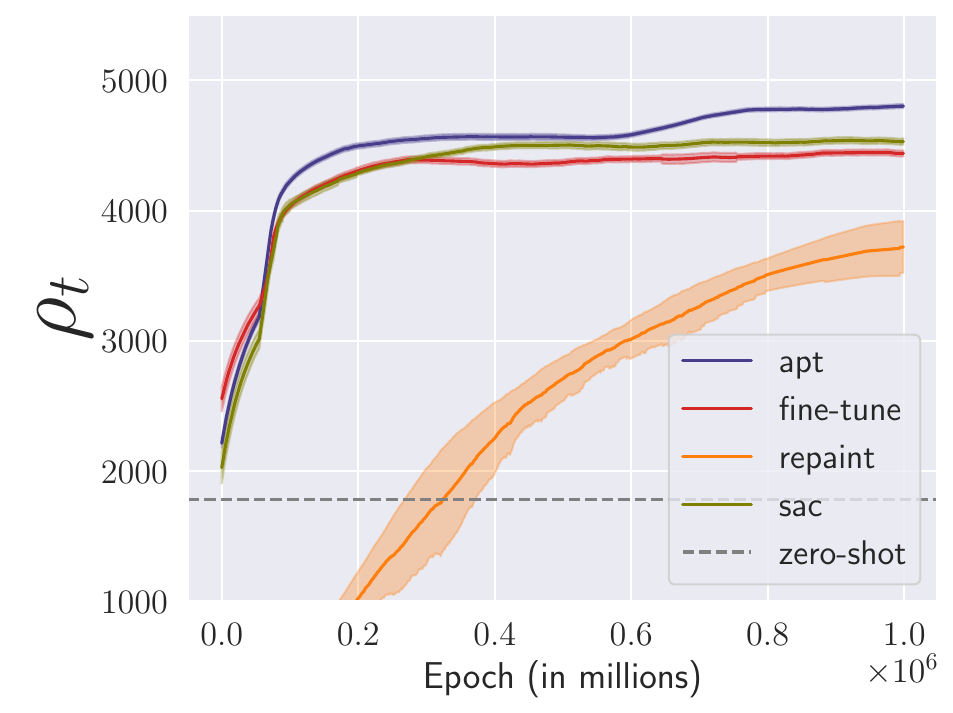} &   
    \includegraphics[width=0.25\textwidth]{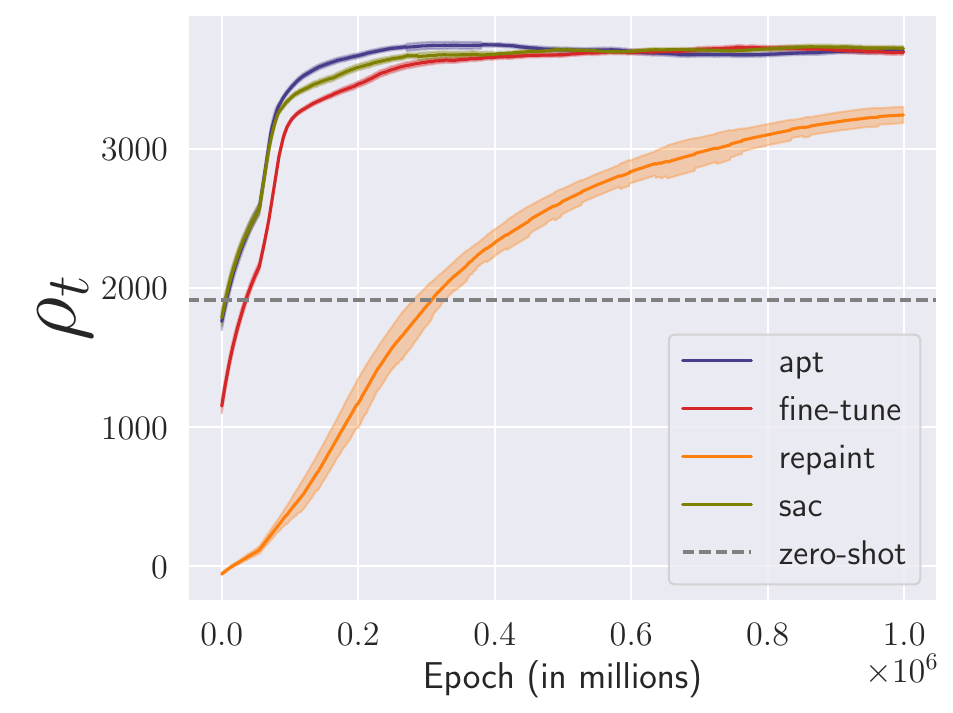} & 
    \includegraphics[width=0.25\textwidth]{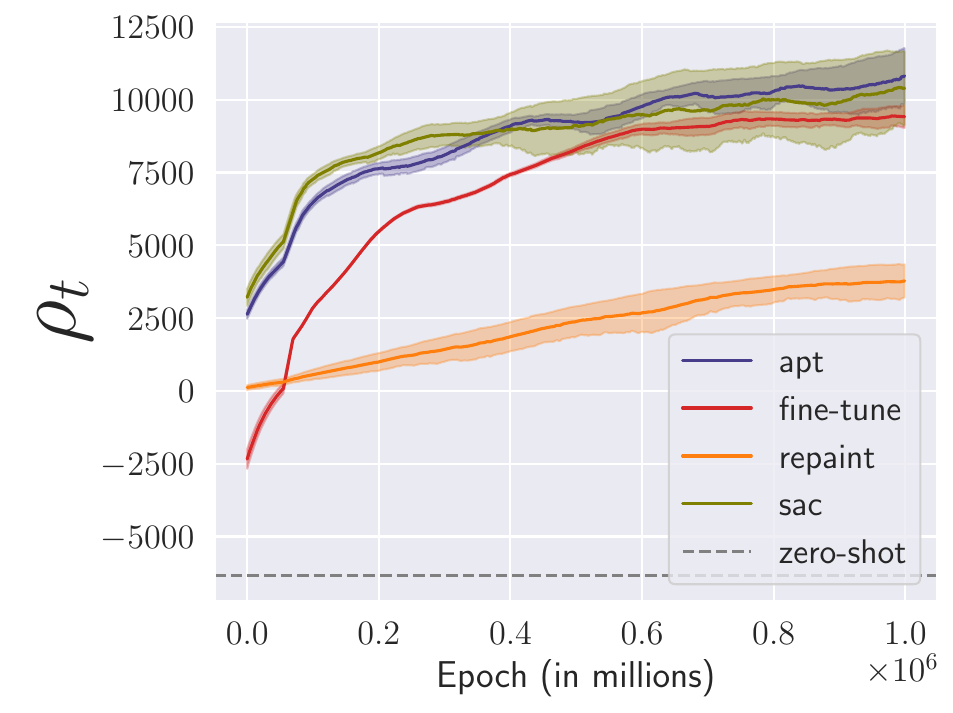} \\  
    \rotatebox{90}{\parbox{3cm}{\centering Ant}} & 
    \includegraphics[width=0.25\textwidth]{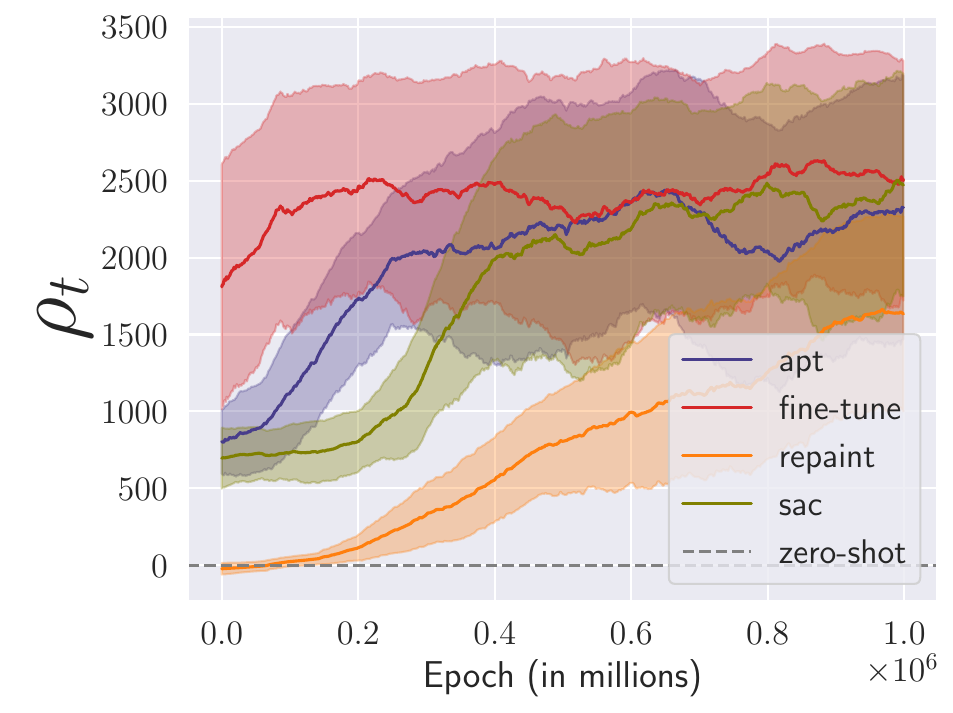} &     
    \includegraphics[width=0.25\textwidth]{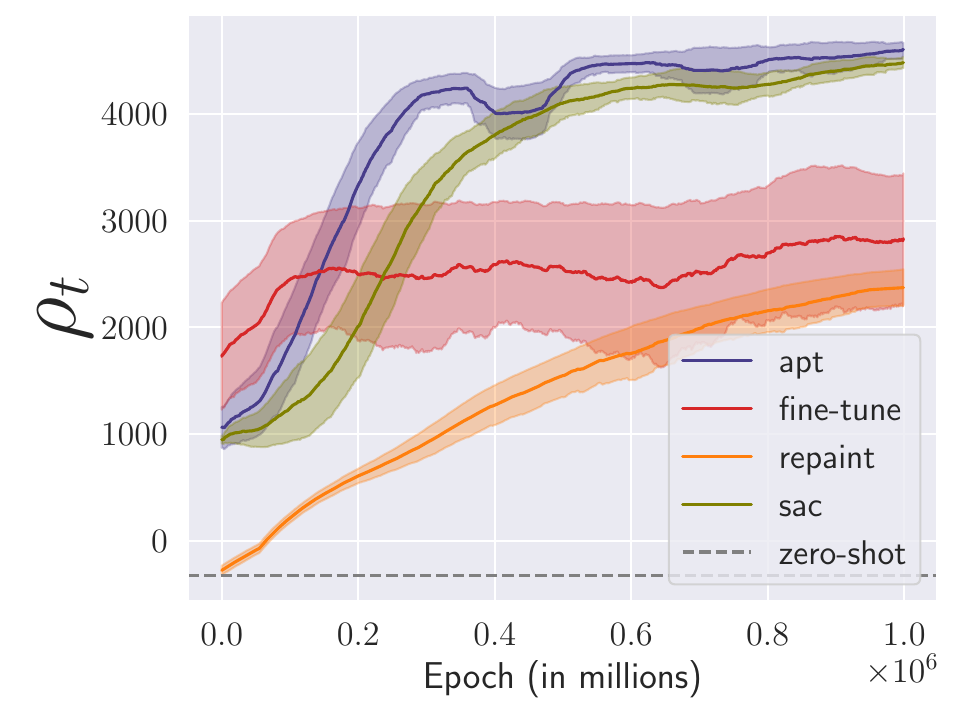} &
    \includegraphics[width=0.25\textwidth]{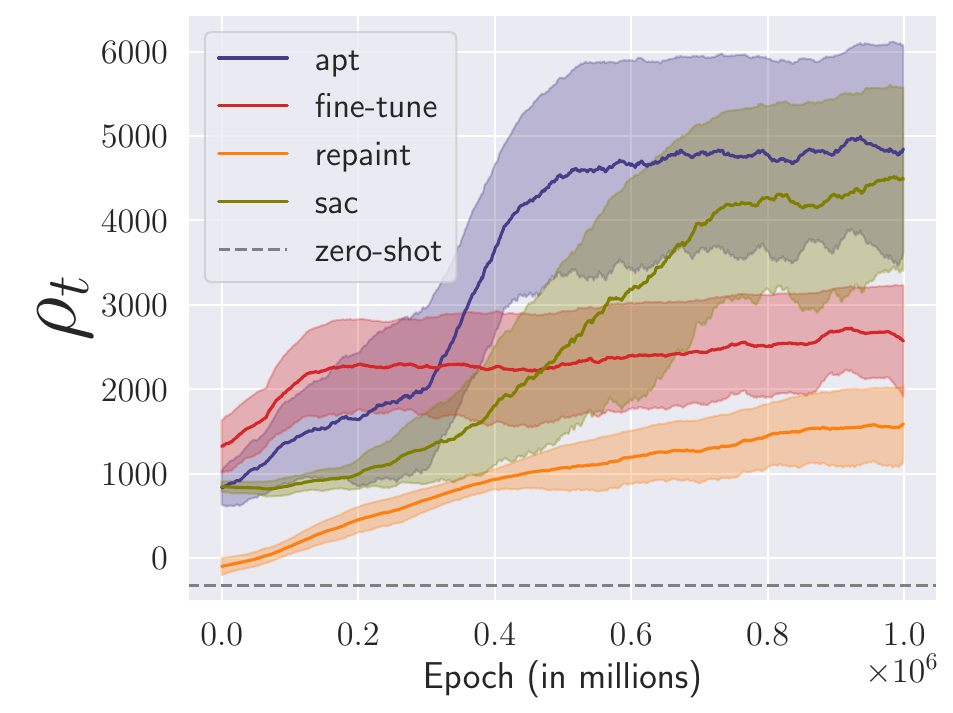} &     
    \includegraphics[width=0.25\textwidth]{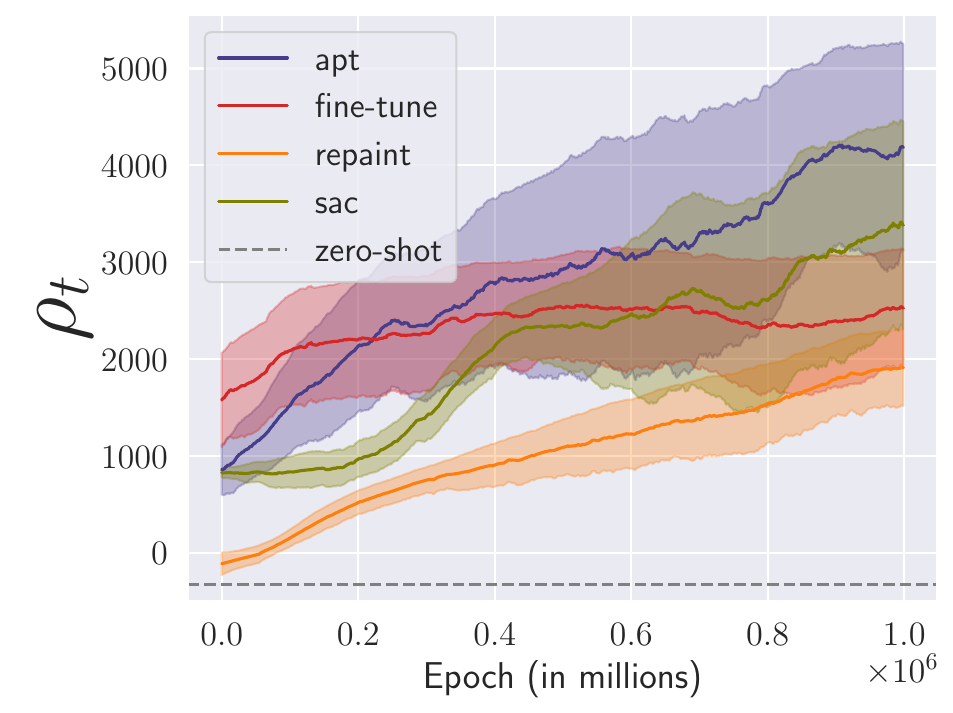} \\
    \rotatebox{90}{\parbox{3cm}{\centering Humanoid}} & 
    \includegraphics[width=0.25\textwidth]{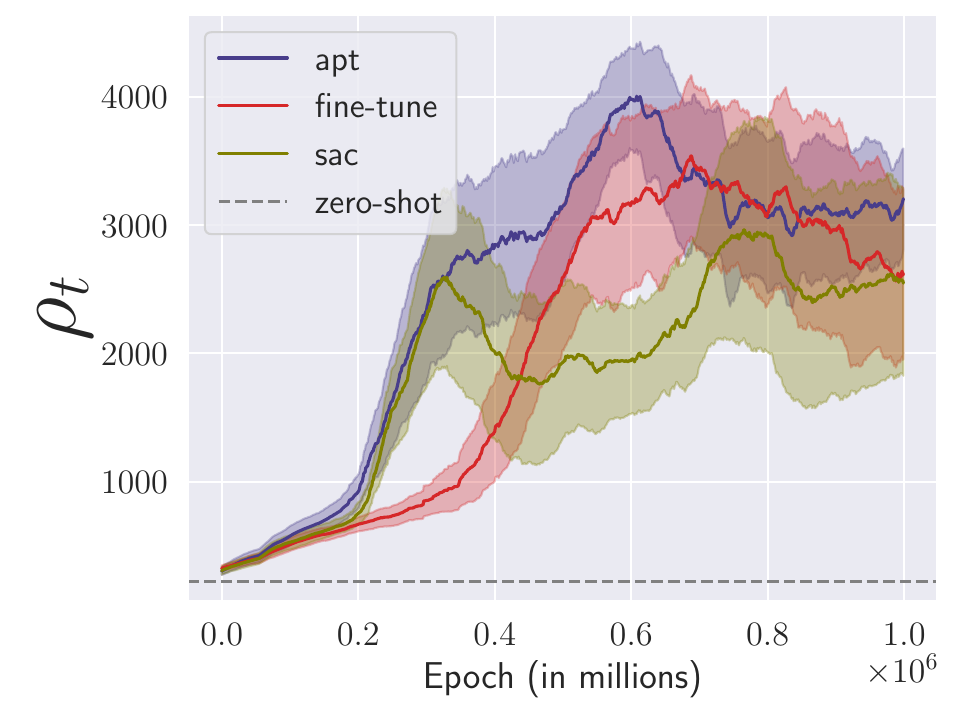} &
    \includegraphics[width=0.25\textwidth]{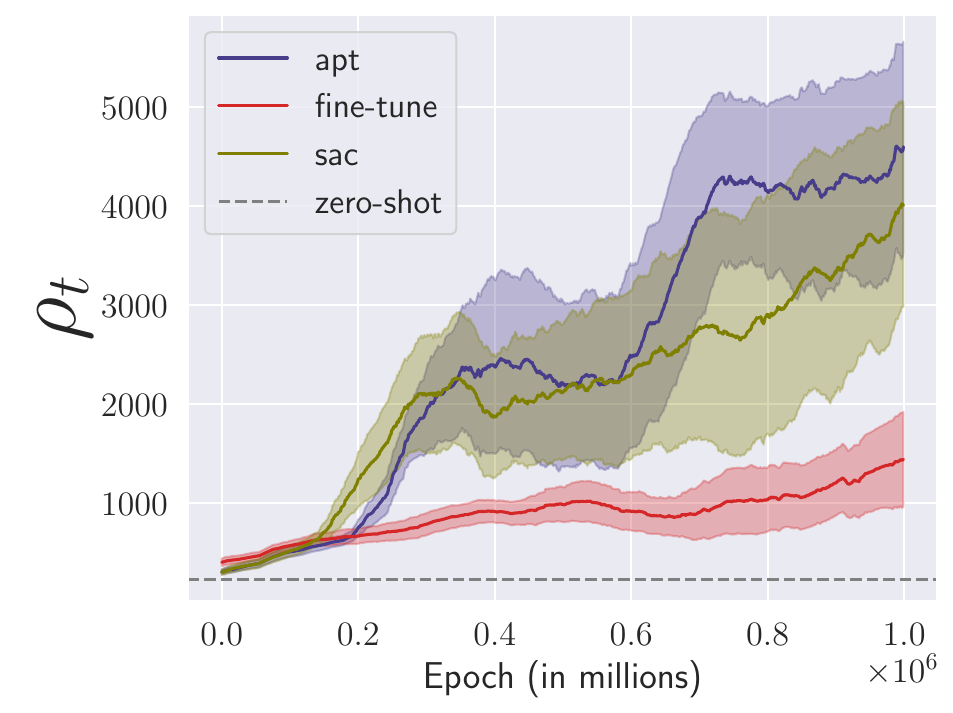} &    
    \includegraphics[width=0.25\textwidth]{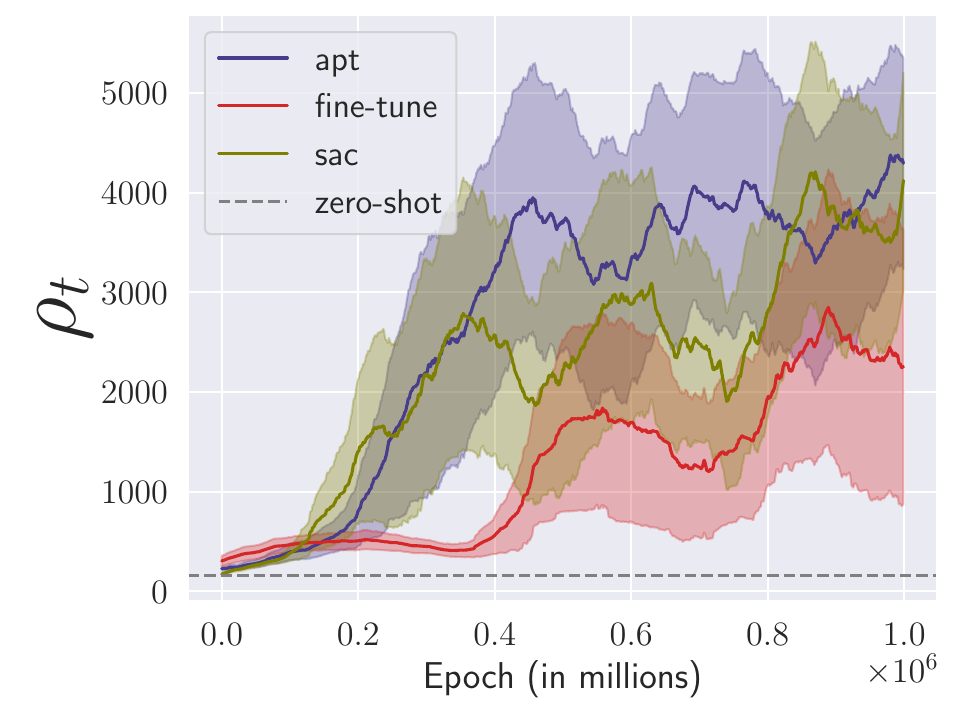} &     
    \includegraphics[width=0.25\textwidth]{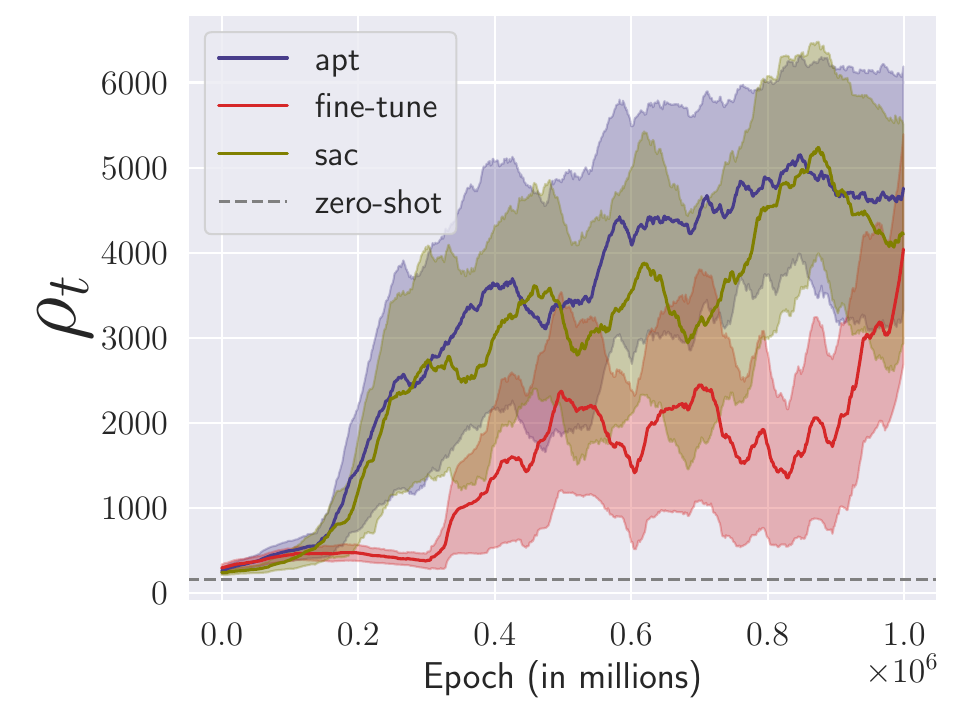} \\     
    & (a) Target task, $\mathcal{T}_1$ & (b) Target task, $\mathcal{T}_2$ & (c) Target task, $\mathcal{T}_3$ & (d) Target task, $\mathcal{T}_4$ 
\end{tabular}
\caption{\textbf{APT-RL transferability, $\Lambda_\text{APT-RL}$:} APT-RL is compared against vanilla SAC (learning from scratch), REPAINT, zero-shot policy, and fine-tuned policy. Average return during the evaluation episode is taken as $\rho_t$, meaning $\rho_t = \mathbb{E}^{\pi^*_{\mathcal{T}_i}}[\sum_{t}r_k]$. We do not show Repaint for the humanoid environment as it fails to solve the tasks. Results are shown with one standard deviation range.}
\label{fig:final_results}
\end{figure}

\begin{figure}[!htb]
\centering
\begin{tabular}{@{}c@{}c@{}}
    \rotatebox{90}{\parbox{3cm}{\centering Half-Cheetah}} & 
        \includegraphics[width=0.75\textwidth]{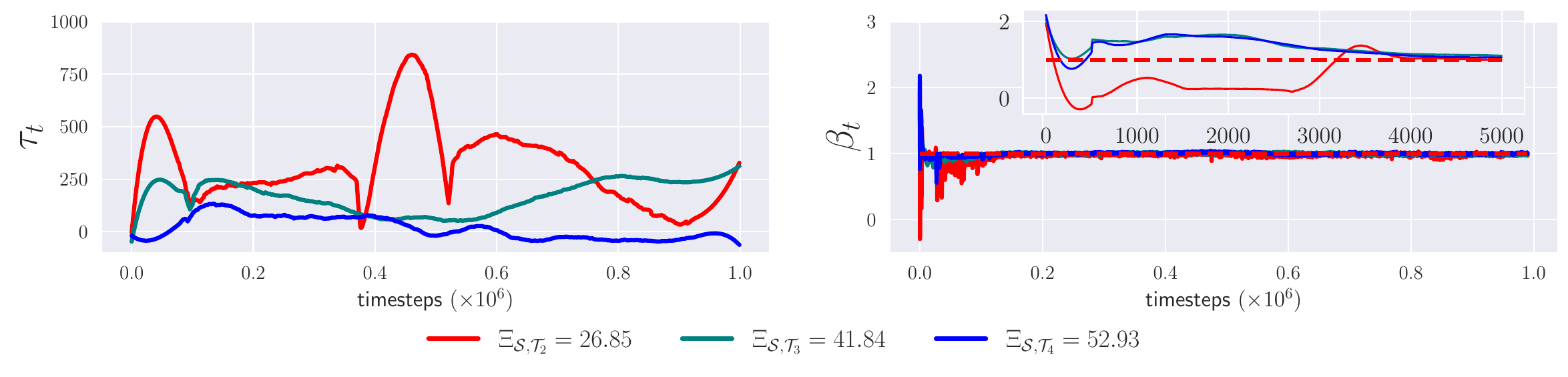}  \\
    \rotatebox{90}{\parbox{3cm}{\centering Ant}} &          \includegraphics[width=0.75\textwidth]{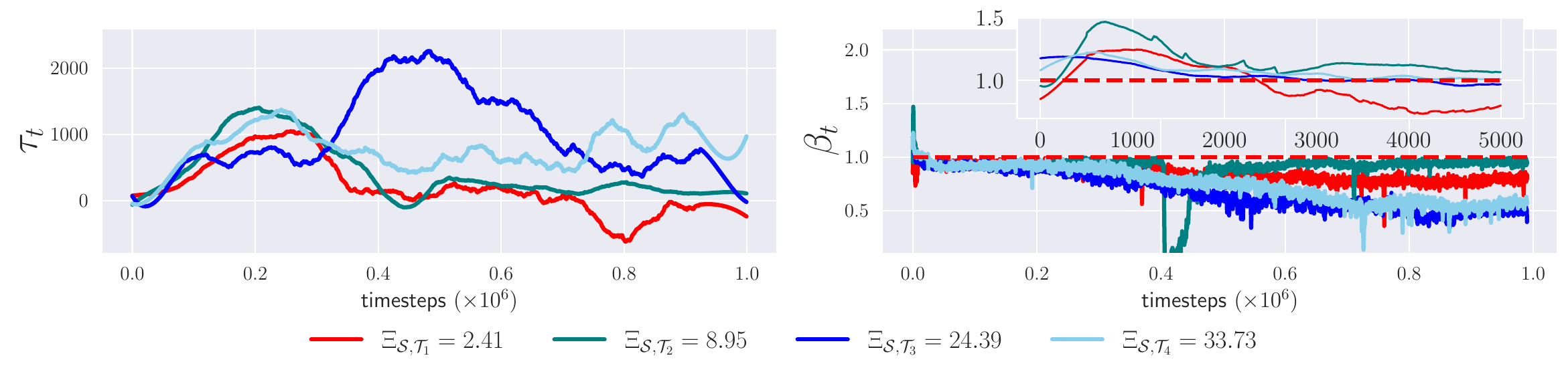} \\
    \rotatebox{90}{\parbox{3cm}{\centering Humanoid}} &          \includegraphics[width=0.75\textwidth]{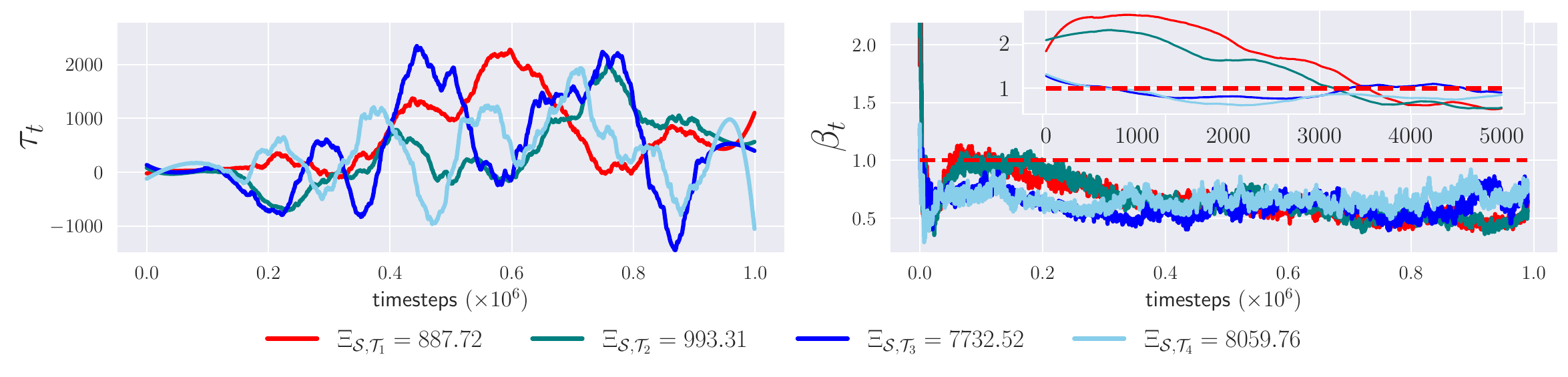} 
\end{tabular}
\caption{Left: Relative transfer performance, $\tau_t$ are shown with corresponding mean similarity scores. Right: Regularization co-efficient, $\beta_t$, is shown for all tasks with corresponding mean similarity scores.}
\label{fig:results_tau}
\end{figure}

Fig. \ref{fig:final_results} (a)-(c) shows the transfer evaluation performance for three target tasks with varying dynamics and Fig. \ref{fig:final_results}(d) shows the performance for one target task with negative reward. In all cases, APT-RL learns faster and also achieves higher average returns than learning from scratch. While fine-tuning the source policy performs better than APT-RL in task $\mathcal{T}_1$, it performs worse than both APT-RL and learning from scratch in rest of the tasks in $\mathcal{T}_2, \mathcal{T}_3, \mathcal{T}_4$. Most importantly, APT-RL performs as good as learning from scratch for target task $\mathcal{T}_4$ with a negative reward. Note that, this is an adversarial source task as the robot needs to learn to run in the opposite direction in the target task. In contrast, the REPAINT algorithm fails to achieve similar evaluation performance and in most cases, obtains evaluation performance lower than learning from scratch using SAC. As REPAINT is a PPO-based on-policy algorithm, this result aligns with the previously reported performance of SAC and PPO algorithms \citep{haarnoja2018soft}. The performance of APT-RL may be explained by the fact that the source optimal policy jumpstarts the target policy. This increase in learning performance, in turn, provides a positive relative transfer measure over time in tasks $\mathcal{T}_1, \mathcal{T}_2, \mathcal{T}_3$ as shown in Fig. \ref{fig:results_tau}. Note that $\tau$ remains higher for the most similar task $\mathcal{T}_1$ and relatively lower for the least similar task $\mathcal{T}_2$. Temperature parameter $\beta$ decreases quickly initially, then increases slightly and stays approximately constant over time. This can be explained by the fact that more weight is put into the regularization loss initially and once the target policy becomes better the effect reduces. As we keep utilizing the target data to update the source policy, the source policy improves over time and provides useful information during the later timesteps. Finally, we observe that the effect of the source policy remains almost constant with respect to task similarity. This makes sense due to the particular change in dynamics of the environment. We anticipate that changing only the damping values make the tasks less adversarial to the source in terms of dynamics.

\begin{figure}
\centering
\begin{tabular}{ccc}
    \includegraphics[width=0.30\textwidth]{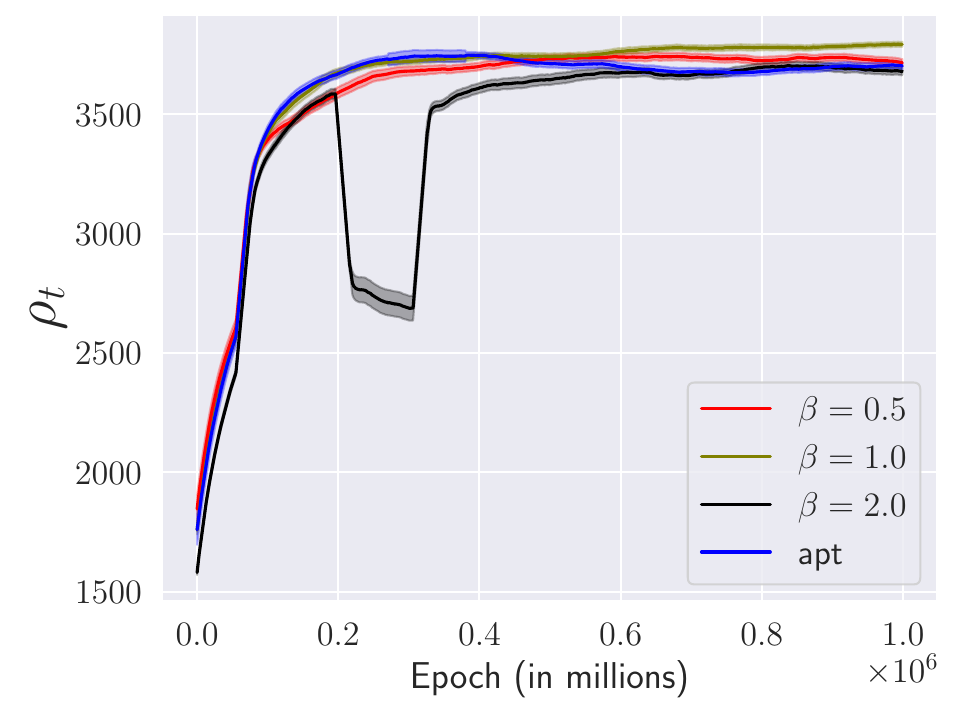} &
   \includegraphics[width=0.30\textwidth]{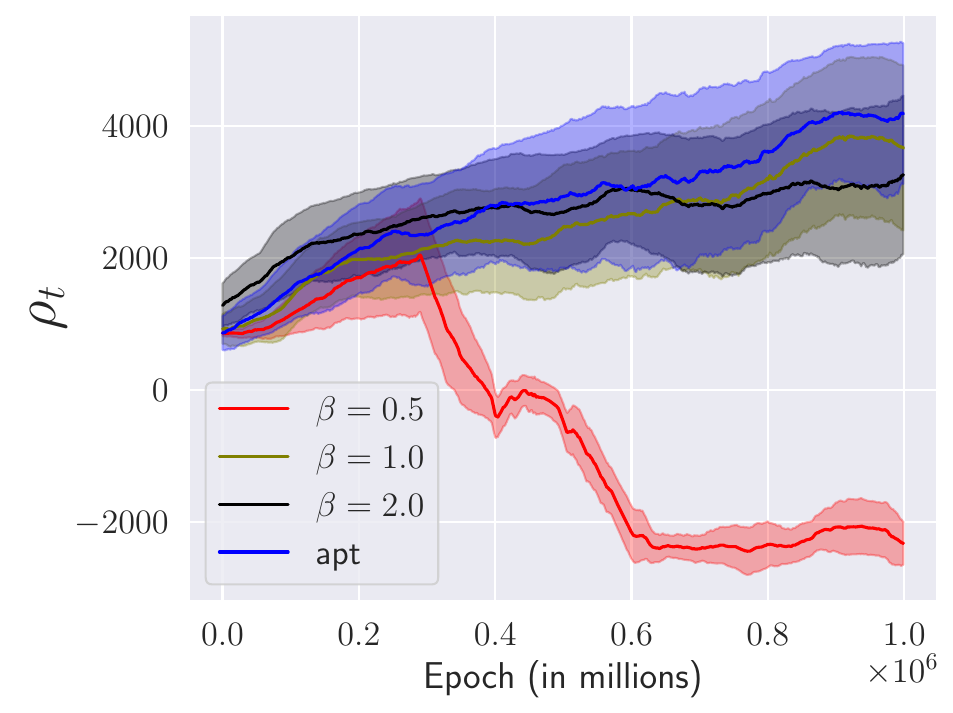} &
    \includegraphics[width=0.30\textwidth]{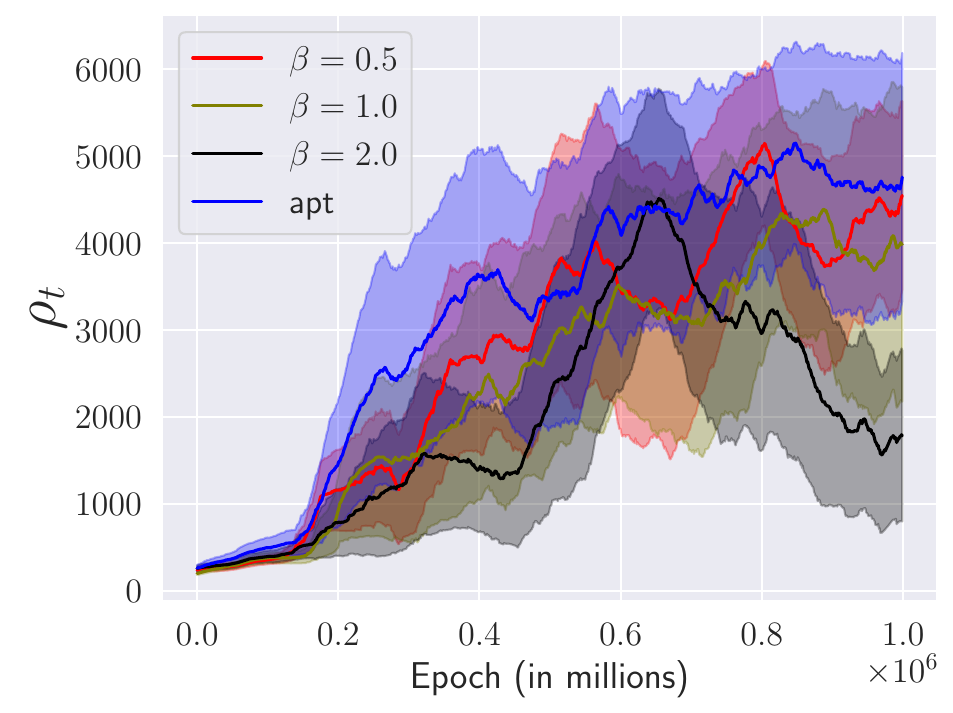} \\ 
    (a) Half-Cheetah & (b) Ant & (c) Humanoid
\end{tabular}
\caption{Ablation study of $\beta$ parameter in APT-RL: manual tuning of hyperparameter $\beta$ is shown against APT-RL in the least similar tasks for all three environments.}
\label{fig:ablationstudy}
\end{figure}

\subsubsection{Ant-v3}
For the ant environment, we observe significant performance gain of APT-RL in the target task against learning from scratch, zero-shot policy transfer, fine-tuning, and the REPAINT algorithm (Fig. \ref{fig:final_results}). For target tasks that are very similar to the source, we observe a fast convergence of the policy in the target. For less similar source and target, APT-RL can even achieve higher learning performance than learning from scratch. This might happen due to the jumpstart of the target policy and also due to the synchronous improvement of the source policy. The latter characteristic of APT-RL accelerates policy updates. Similar to the half-cheetah environment, we observe that the temperature parameter decreases with timesteps and decreases more when task similarity is lower (Fig. \ref{fig:results_tau}). In all of these examples, APT-RL outperforms the REPAINT algorithm. 

\subsection{Humanoid-v3}
For the humanoid environment, APT-RL outperforms all the baselines with significant initial performance gain (Fig. \ref{fig:final_results}). For the final target task, $\mathcal{T}_4$, APT-RL performs similarly to learning from scratch. We anticipate that this behavior is mainly due to the difficulty of the task. The change in the dynamics of the environment make the target an adversarial task which is relatively more difficult to solve than the rest of the tasks. This can be also supported by the fact that $\mathcal{T}_4$ is the least similar task among all of the target tasks. Similar to the ant and half-cheetah environments, we observe that the temperature parameter decreases with less task similarity (Fig. \ref{fig:results_tau}). We do not show results from the REPAINT algorithm as it fails to solve even the source task.  

Finally, we show an ablation study on the effect of the temperature parameter, $\beta$, in Fig. \ref{fig:ablationstudy}. Notice that APT-RL outperforms manual choice of $\beta$ in all tasks except the halfcheetah environment where $\beta = 1.0$ performs slightly better than APT-RL. Interestingly, Fig. \ref{fig:results_tau} shows that APT-RL also converges to $\beta = 1.0$ without any manual choice. We argue that it shows further evidence on the strength of APT-RL where we do not need to manually choose the temperature parameter.

\section{Limitations} 
The task similarity algorithm presented in section \ref{sec:task_sim} uses the random policy to learn models for the dynamics and the reward. While using a random policy for model learning is fairly common in the literature \cite{zhang2018decoupling, moerland2023model}, a key challenge is to learn a reasonable model for complex tasks. We anticipate that the task similarity algorithm might not be effective in environments where a random policy cannot efficiently capture the underlying complexities of the dynamics and the reward. We would also like to focus on the limitations of APT-RL. One of the key challenges of transfer RL is to identify useful source tasks and reduce the impact of adversarial sources. While APT-RL shows strong performance in most of the cases, as shown in the Fig. \ref{fig:final_results}(a) for the 'Half-Cheetah-V3' environment, simple transfer approaches such as fine-tuning the source policy is more convenient. This behavior can be explained by the close similarity of the source to the target. Therefore, identifying measures of when to opt for more advanced transfer algorithms such as APT-RL remains an interesting challenge.

\section{Conclusion}

In this paper, we proposed the APT-RL algorithm to transfer knowledge from a source task in an off-policy fashion. Through advantage-based regularization, our algorithm does not require any heuristic or manual fine-tuning of the objective function. We also introduced a new relative transfer performance measure, which can help evaluate and compare transfer learning approaches in RL. We also provided a simple, theoretically-backed algorithm to calculate task similarity, and demonstrated the alignment of our proposed transfer performance measure with source and target task similarities. We demonstrated the effectiveness of APT-RL in continuous control tasks and showed its superior performance against benchmark transfer RL algorithms. Future directions may include considering similar concepts for multi-task transfer learning scenarios, as well as benchmarking the performance of various transfer learning algorithms with the help of the transferability measures introduced in this paper.

\bibliography{main}
\bibliographystyle{tmlr}

\newpage
\appendix
\section{Proof of Theorem 1}\label{app:thm1}
\begin{customthm}{1}\label{thm:relative-transfer-repeated}{\textbf{(Relative transfer performance and policy improvement)}} Consider $\rho_t^i = \mathbb{E}^{\pi_{i, t}}\left[\sum_{k=0}^H r_k|\mathbf{s}_0\right]$ for policy $\pi_i$ and $\rho_t^b = \mathbb{E}^{\pi_{b, t}}\left[\sum_{k=0}^H r_k|\mathbf{s}_0\right]$ for policy $\pi_b$, where $\mathbf{s}_0$ is the starting state and each policy is executed for $H$ timesteps, then the learned policy, $\pi_{i, t}$ using algorithm $i$, in the target at episode $t$ is at least as good as the source optimal policy, $\pi_{b, t}$ if $\tau_t \geq 0$.  
\end{customthm}

\begin{proof}
Let us consider $\tau_t = \mathbb{E}_{\pi_{i, t}}[G|\mathbf{s}_0] - \mathbb{E}_{\pi_{\mathbf{s}, t}}[G|\mathbf{s}_0]$ where $\pi_i$ is the current policy in the task $i$ and $\pi_b$ is the optimal base policy and $\mathbf{s}_0$ is the initial state drawn from the same distribution. Using this definition of $\tau$, we can write the following for any starting state $s_0$: 
\begin{align*}
    \tau_t \geq 0 \\ 
    \Rightarrow \mathbb{E}_{\pi_{i, t}}[G|\mathbf{s}_0] - \mathbb{E}_{\pi_{b, t}}[G|\mathbf{s}_0] \geq 0\\ 
    \Rightarrow V_{\pi_{i, t}}(\mathbf{s}_0) - V_{\pi_{b, t}}[(\mathbf{s}_0) \geq 0\\ 
    \Rightarrow V_{\pi_{i, t}}(\mathbf{s}_0) \geq V_{\pi_{b, t}}[(\mathbf{s}_0)
\end{align*}
Following the policy improvement theorem, we can say that $\pi_{i, t}$ is at least as good as $\pi_{b, t}$. 
\end{proof}

\section{Proof of Theorem 2}\label{app:thm2}
\begin{customthm}{2}\label{thm:bound-repeated}{\textbf{(Action-value bound between fixed-domain environments)}} If $\pi^*_\mathcal{S}$ and $\mathcal{\pi^*_\mathcal{T}}$ are the optimal policies in the MDPs $\mathcal{M}_\mathcal{S} = \langle \mathcal{X}, \mathcal{A}, \mathcal{R}_\mathcal{S}, \mathcal{P}_\mathcal{S}\rangle$ and $\mathcal{M}_\mathcal{T} = \langle \mathcal{X}, \mathcal{A}, \mathcal{R}_\mathcal{T}, \mathcal{P}_\mathcal{T}\rangle$ respectively, then the corresponding action-value functions can be upper bounded by 
\begin{equation}
    ||\mathbf{Q}^{\pi^*_\mathcal{T}}_\mathcal{T} -  \mathbf{Q}^{\pi^*_\mathcal{S}}_\mathcal{T}||_\infty \leq \frac{2\delta^r_{\mathcal{S}\mathcal{T}}}{1-\gamma} + \frac{2\gamma \delta^{TV}_{\mathcal{S}\mathcal{T}}(R_{max, \mathcal{S}} + R_{max, \mathcal{T}})}{(1-\gamma)^2}
    \label{eq:bound-new-repeated}
\end{equation}

where $\delta^r_{\mathcal{S}\mathcal{T}} = ||\mathcal{R}_\mathcal{S}(\mathbf{s}, \mathbf{a})- \mathcal{R}_\mathcal{T}(\mathbf{s}, \mathbf{a}))||_\infty$, $\delta^{TV}_{\mathcal{S} \mathcal{T}}$ is the total variation distance between $\mathcal{P}_\mathcal{S}$ and $\mathcal{P}_\mathcal{T}$, $\gamma$ is the discount factor and $R_{max, \mathcal{S}} = ||\mathcal{R}_\mathcal{S}(\mathbf{s}, \mathbf{a})||_\infty$, $R_{max, \mathcal{T}} = ||\mathcal{R}_\mathcal{T}(\mathbf{s}, \mathbf{a})||_\infty$.
\end{customthm}

\begin{proof}
Let us consider the following notations for simplicity, $Q^{\pi^*_i}_i(s, a) \equiv Q^i_i(s, a), Q^{\pi^*_j}_j(s, a) \equiv Q^j_j(s, a), Q^{\pi^*_j}_i(s, a) \equiv Q^j_i(s, a)$. Now we can write the following, 
\begin{align*}
    |Q_i^i(s, a) - Q^j_i(s, a)| =& |Q_i^i(s, a) - Q_j^j(s, a) + Q_j^j(s, a) - Q^j_i(s, a)|\\
    \leq& \underbrace{|Q_i^i(s, a) - Q_j^j(s, a)|}_{(a)} + \underbrace{|Q_j^j(s, a) - Q^j_i(s, a)|}_{(b)}
\end{align*}
Our strategy is to find bounds for (a) and (b) separately and then combine them to get the final bound. 

\subsection*{(a)}
\begin{align*}
    & |Q_i^i(s, a) - Q^j_j(s, a)|\\ 
                    =& \left|r_i(s, a) + \gamma \sum_{s'}p_i(s'|s, a) \max_b Q_i^i(s', b) - r_j(s, a) - \gamma \sum_{s'}p_j(s'|s, a) \max_b Q_j^j(s', b)\right|\\
                    \leq& |r_i(s, a) - r_j(s, a)| + \gamma \left|\sum_{s'}\underbrace{p_i(s'|s, a)}_{p_i} \max_b Q_i^i(s', b) -  \sum_{s'}\underbrace{p_j(s'|s, a)}_{p_j}\max_b Q_j^j(s', b)\right|\\ 
                    =& |r_i(s, a) - r_j(s, a)| + \gamma \left|\sum_{s'} p_i\max_b Q_i^i(s', b) - p_j\max_b Q_i^i(s', b) + p_j\max_b Q_i^i(s', b) -  p_j \max_b Q_j^j(s', b)\right|\\
                    =& |r_i(s, a) - r_j(s, a)| + \gamma \left|\sum_{s'} ( p_i - p_j)\max_bQ^i_i(s', b) + p_j\left(\max_b Q_i^i(s', b) -  \max_b Q_j^j(s', b) \right)\right|\\
                    \leq& |r_i(s, a) - r_j(s, a)| + \gamma \sum_{s'}\left|( p_i - p_j)\max_bQ^i_i(s', b) + p_j\left(\max_b Q_i^i(s', b) -  \max_b Q_j^j(s', b)\right) \right|\\
                    \leq& \underbrace{|r_i(s, a) - r_j(s, a)|}_{T_1} + \gamma \underbrace{\sum_{s'}\left|( p_i - p_j)\max_bQ^i_i(s', b)\right|}_{T_2} + \gamma\underbrace{\sum_{s'}\left| p_j\left(\max_b Q_i^i(s', b) -  \max_b Q_j^j(s', b)\right) \right|}_{T_3}
\end{align*}

Let us consider each term of the above equation individually to calculate the bound. For convenience, we can consider the following vector notations,
\begin{align*}
\mathbf{Q}_i^i &= \begin{bmatrix} \max_bQ_i^i(s', b), \dots\end{bmatrix}^T  \quad \quad \forall s'\in \mathcal{S}\\ 
\mathbf{Q}_j^j &= \begin{bmatrix} \max_bQ_j^j(s', b), \dots\end{bmatrix}^T \quad \quad\forall s'\in \mathcal{S}\\  
\mathbf{R}_i &= \begin{bmatrix} r_i(s, a), \dots\end{bmatrix}^T \quad \quad \quad \quad \quad \forall s\in \mathcal{S}, a \in \mathcal{A}\\
\mathbf{R}_j &= \begin{bmatrix} r_j(s, a), \dots\end{bmatrix}^T  \quad \quad \quad \quad \quad \forall s\in \mathcal{S}, a \in \mathcal{A}\\ 
\mathbf{P}_i &= \begin{bmatrix} p_i(s'|s, a), \dots\end{bmatrix}^T  \quad \quad \quad \quad \forall s'\in \mathcal{S}, a \in \mathcal{A}\\ 
\mathbf{P}_j &= \begin{bmatrix} p_j(s'|s, a), \dots\end{bmatrix}^T  \quad \quad \quad \quad \forall s'\in \mathcal{S}, a \in \mathcal{A}
\end{align*}

Using these notations, we can rewrite $T_2$ as the following, 
\begin{align*}
    & \sum_{s'}\left|( p_i - p_j)\max_bQ^i_i(s', b)\right| \\
    &= \left\Vert\left(\mathbf{P}_i - \mathbf{P}_j\right) \cdot \mathbf{Q}^i_i\right\Vert_1 \\  
    &\leq \left\Vert \mathbf{P}_i - \mathbf{P}_j \right\Vert_1 \left\Vert \mathbf{Q}_i^i \right\Vert_\infty  \quad \quad \text{using H\"{o}lder's inequality}, ||fg||_1 \leq ||f||_p ||g||_q \text{ where } \frac{1}{p} + \frac{1}{q} = 1  \\ 
    &= 2 \delta^{TV}_{ij}\left\Vert \mathbf{Q}_i^i \right\Vert_\infty  \quad \quad \text{where  } \delta_p^{ij} \text{ is the total variation distance, } \delta^{T}_{ij} = D_{TV} (\mathbf{P}_i, \mathbf{P}_j)
\end{align*}
Similarly, we can write for $T_3$, 
\begin{align*}
    & \sum_{s'}\left|p_j\left(\max_bQ^i_i(s', b) - \max_bQ^j_j(s', b)\right)\right| \\
    &\leq \left\Vert \mathbf{P}_j \right\Vert_1 \left\Vert \mathbf{Q}_i^i - \mathbf{Q}_j^j \right\Vert_\infty\\ 
    & = \left\Vert \mathbf{Q}_i^i - \mathbf{Q}_j^j \right\Vert_\infty \quad \quad \text{because } \left\Vert \mathbf{P}_j \right\Vert_1 = 1
\end{align*}
Thus we can write the following, 
\begin{align*}
|Q_i^i(s, a) - Q^j_j(s, a)| \leq \Vert \mathbf{R}_i - \mathbf{R}_j\Vert_\infty + 2 \gamma\delta^{TV}_{ij}\mathbf{Q}_i^i + \gamma\left\Vert \mathbf{Q}_i^i - \mathbf{Q}_j^j \right\Vert_\infty
\end{align*}

Because this is true for all $a\in \mathcal{S}, a \in \mathcal{A}$, we can write the following, 
\begin{align*}
    ||\mathbf{Q}_i^i -  \mathbf{Q}_j^j||_\infty 
                \leq& \Vert \mathbf{R}_i - \mathbf{R}_j\Vert_\infty + 2 \gamma\delta^{TV}_{ij}\mathbf{Q}_i^i + \gamma\left\Vert \mathbf{Q}_i^i - \mathbf{Q}_j^j \right\Vert_\infty\\ 
    \Rightarrow ||\mathbf{Q}_i^i -  \mathbf{Q}_j^j||_\infty &\leq \frac{\delta^r_{ij}}{1-\gamma} + \frac{2\gamma\delta^{TV}_{ij}}{1-\gamma} ||\mathbf{Q}^i_i||_\infty\\
    \Rightarrow ||\mathbf{Q}_i^i -  \mathbf{Q}_j^j||_\infty &\leq \frac{\delta^r_{ij}}{1-\gamma} + \frac{2\gamma R_{max, i}\delta^{TV}_{ij}}{(1-\gamma)^2}
\end{align*}

\subsection*{(b)}
\begin{align*}
    & |Q_j^js, a) - Q^j_i(s, a)|\\ 
    &= \left|r_j(s, a) + \gamma \sum_{s'}p_j(s'|s, a) \max_b Q_j^j(s', \pi_j(s)) - r_i(s, a) - \gamma \sum_{s'}p_j(s'|s, a) \max_b Q_j^i(s', \pi_j(s))\right|\\
    &\leq \left|r_j(s, a) - r_i(s, a) \right| + \gamma \left| \sum_{s'}p_j(s'|s, a) \max_b Q_j^j(s', \pi_j(s)) - \sum_{s'}p_i(s'|s, \pi_i(s)) \max_b Q_j^i(s', \pi_j(s)) \right|\\
    &\leq \left|r_j(s, a) - r_i(s, a) \right| + \gamma \sum_{s'}\left| p_j \max_b Q_j^j(s', \pi_j(s)) - p_i \max_b Q_j^i(s', \pi_j(s)) \right|\\
    &\leq \left|r_j(s, a) - r_i(s, a) \right| + \gamma \sum_{s'}\left| p_j \max_b Q_j^j(s', \pi_j(s)) - p_i \max_b Q_j^j(s', \pi_j(s)) + p_i \max_b Q_j^j(s', \pi_j(s)) - p_i \max_b Q_j^i(s', \pi_j(s)) \right|\\
    &= \left|r_j(s, a) - r_i(s, a) \right| + \gamma \sum_{s'}\left| \left(p_j - p_i\right) \max_b Q_j^j(s', \pi_j(s)) + p_i \left(\max_b Q_j^i(s', \pi_j(s)) - \max_b Q_j^i(s', \pi_j(s))\right) \right|\\
    &\leq \left|r_j(s, a) - r_i(s, a) \right| + \gamma \sum_{s'}\left| \left(p_j - p_i\right) \max_b Q_j^j(s', \pi_j(s))\right| + \gamma \sum_{s'}\left| p_i \left(\max_b Q_j^i(s', \pi_j(s)) - \max_b Q_j^i(s', \pi_j(s))\right) \right|\\
    &\leq \left\Vert \mathbf{R}_i - \mathbf{R}_j\right\Vert + \gamma \left\Vert P_i - P_j\right\Vert_1 \left\Vert\mathbf{Q}_j^j\right\Vert_\infty + \gamma\left\Vert \mathbf{P}_i\right\Vert_1 \left\Vert\mathbf{Q}_j^j - \mathbf{Q}_i^j\right\Vert_\infty \\
    &\leq \delta^r_{ij} + \frac{2\gamma \delta^{TV}_{ij}R_{max, j}}{1-\gamma} + \gamma\left\Vert\mathbf{Q}_j^j - \mathbf{Q}_i^j\right\Vert_\infty \\
\end{align*}

Because this is true for all $a\in \mathcal{S}, a \in \mathcal{A}$, we can write the following, 
\begin{align*}
    ||\mathbf{Q}_j^j -  \mathbf{Q}_j^i||_\infty 
    &\leq \frac{\delta^r_{ij}}{1-\gamma} + \frac{2\gamma \delta^{TV}_{ij}R_{max, j}}{(1-\gamma)^2}\\
\end{align*}

Finally, we can combine (a) and (b) to write the following, 
\begin{align*}
    ||\mathbf{Q}_i^i -  \mathbf{Q}_i^j||_\infty &\leq \frac{2\delta^r_{ij}}{1-\gamma} + \frac{2\gamma \delta^{TV}_{ij}(R_{max, i} + R_{max, j})}{(1-\gamma)^2}\\
\end{align*}
\end{proof}

\section{Experiment details}\label{app:exp_details}
\begin{table}
\begin{center}
    \begin{tabular}{c c c c c c c }
    \hline 
     environment &  change type & source & Target 1 & Target 2& Target 3& Target 4\\ 
     & & ($\mathcal{S}$)& ($\mathcal{T}_1$)& $(\mathcal{T}_2)$&$(\mathcal{T}_3)$ & $(\mathcal{T}_4)$ \\
     \hline 
     \multirow{1}{*}{HalfCheetah-v3}  &  \begin{tabular}{c}bthigh damping\\bshin  damping\\ bfoot  damping\\ fthigh  damping\\ fshin  damping\\ffoot  damping\end{tabular}
     & \begin{tabular}{c} 6.0\\4.5\\3.0\\4.5\\3.0\\1.5\\ \end{tabular}
     & \begin{tabular}{c} 9.0\\6.0\\3.0\\9\\6.0\\3.0\\ \end{tabular}
     & \begin{tabular}{c} 12.0\\9.0\\6.0\\12.0\\9.0\\6.0\\ \end{tabular}
     & \begin{tabular}{c} 15.0\\12.0\\9.0\\15.0\\12.0\\9.0\\ \end{tabular}
     & \begin{tabular}{c} 18.0\\15.0\\12.0\\18.0\\15.0\\12.0\\ \end{tabular}\\ 
    \hline
    \multirow{1}{*}{Ant-v3}  &  \begin{tabular}{c}fright upper length\\fright lower length\\ fleft upper length \\ fleft lower length \\ bright upper length \\bright lower length\\bleft upper length \\ bleft lower length \end{tabular}
     & \begin{tabular}{c} 0.2\\0.4\\0.2\\0.4\\0.2\\0.4\\0.2\\0.4 \end{tabular}
     & \begin{tabular}{c} 0.2\\0.5\\0.2\\0.5\\0.2\\0.5\\0.2\\0.5\end{tabular}
     & \begin{tabular}{c} 0.2\\0.2\\0.2\\0.2\\0.2\\0.2\\0.2\\0.2\end{tabular}
     & \begin{tabular}{c} 0.2\\0.3\\0.2\\0.3\\0.2\\0.5\\0.2\\0.5\end{tabular}
     & \begin{tabular}{c} 0.2\\0.3\\0.2\\0.3\\0.2\\0.3\\0.2\\0.5\end{tabular}\\ 
    \hline    
    \multirow{1}{*}{Humanoid-v3}  &  \begin{tabular}{c}right shin length\\right upper arm\\ left upper arm\\ right shin size \\ left shin size \\ right foot size \\ left foot size\\ right lower arm size\\ left lower arm size\\ right hand size\\left hand size \end{tabular}
     & \begin{tabular}{c} 0.30\\0.277\\0.277\\0.049\\0.049\\0.075\\0.075\\ 0.031\\0.031\\0.04\\0.04\end{tabular}
     & \begin{tabular}{c} 0.05\\0.277\\0.277\\0.049\\0.049\\0.075\\0.075\\0.031\\0.031\\0.04\\0.04\end{tabular}
     & \begin{tabular}{c} 0.05\\0.139\\0.139\\0.049\\0.049\\0.075\\0.075\\0.031\\0.031\\0.04\\0.04\end{tabular}
     & \begin{tabular}{c} 0.30\\0.277\\0.277\\0.01\\0.01\\0.01\\0.01\\0.031\\0.031\\0.04\\0.04\end{tabular}
     & \begin{tabular}{c} 0.30\\0.277\\0.277\\0.01\\0.01\\0.01\\0.01\\0.01\\0.01\\0.01\\0.01\end{tabular}
     \\
    \hline    
    \end{tabular}
\end{center}
    \caption{Target task specifications}
    \label{table:exp}
\end{table}

\noindent \textbf{HalfCheetah-v3:} This is a complex continuous control task of a 2D cat-like robot where the objective is to apply torque on the joints to make it run as fast as possible. The observation space is $17$-dimensional and the action space, $a \in \mathbb{R}^6 \forall a \in \mathcal{A}$. Each action is a torque applied to one of the front or back rotors of the robot and can take a value between $[-1.0, 1.0]$. We make two types of perturbations to create target tasks; reward variation and dynamics variation. For the reward variation, the source environment uses a forward reward of $+1$, and the four target environments have a forward reward $r = [-2, -1, 1, 2]$. Note that a negative forward reward is a target task where the robot needs to run in the opposite direction than the source. For this type of example, the source acts as an adversarial task and the goal is to learn at least as good as learning from scratch. Next, we consider target tasks with varying dynamics. The source environment has the standard gym values for damping and the four target environments have different values of damping increased gradually in each task. The least similar task has the highest damping values in the joints. 

\noindent \textbf{Ant-v3:} This is also a high dimensional continuous control task where the goal is to make an ant-robot move in the forward direction by applying torques on the hinges that connect each leg and torso of the robot. The observation space is $27$-dimensional and the action space, $a\in \mathbb{R}^8 \forall a\in\mathcal{A}$ where each action is a torque applied at the hinge joints with a value between $[-1.0, 1.0]$. The source environment has the standard gym robot and the four target environments have varied dynamics by changing the leg lengths of the robot. Representative figures of these dynamics can be found in the appendix.

\noindent \textbf{Humanoid-v3:} This is a high-dimensional continuous control task where the goal is to make a humanoid robot move in the forward direction by applying torques on the hinge joints. The observation space is $376$-dimensional and the action space, $a\in \mathbb{R}^{17} \forall a\in\mathcal{A}$ where each action is a torque applied at the hinge joints with a value between $[-0.4, 0.4]$. The source environment has the standard gym robot and the four target environments have varied dynamics by changing the hand and leg lengths as well as sizes. 

\begin{figure}
\centering
\begin{tabular}{cc}
    \includegraphics[width=0.40\textwidth]{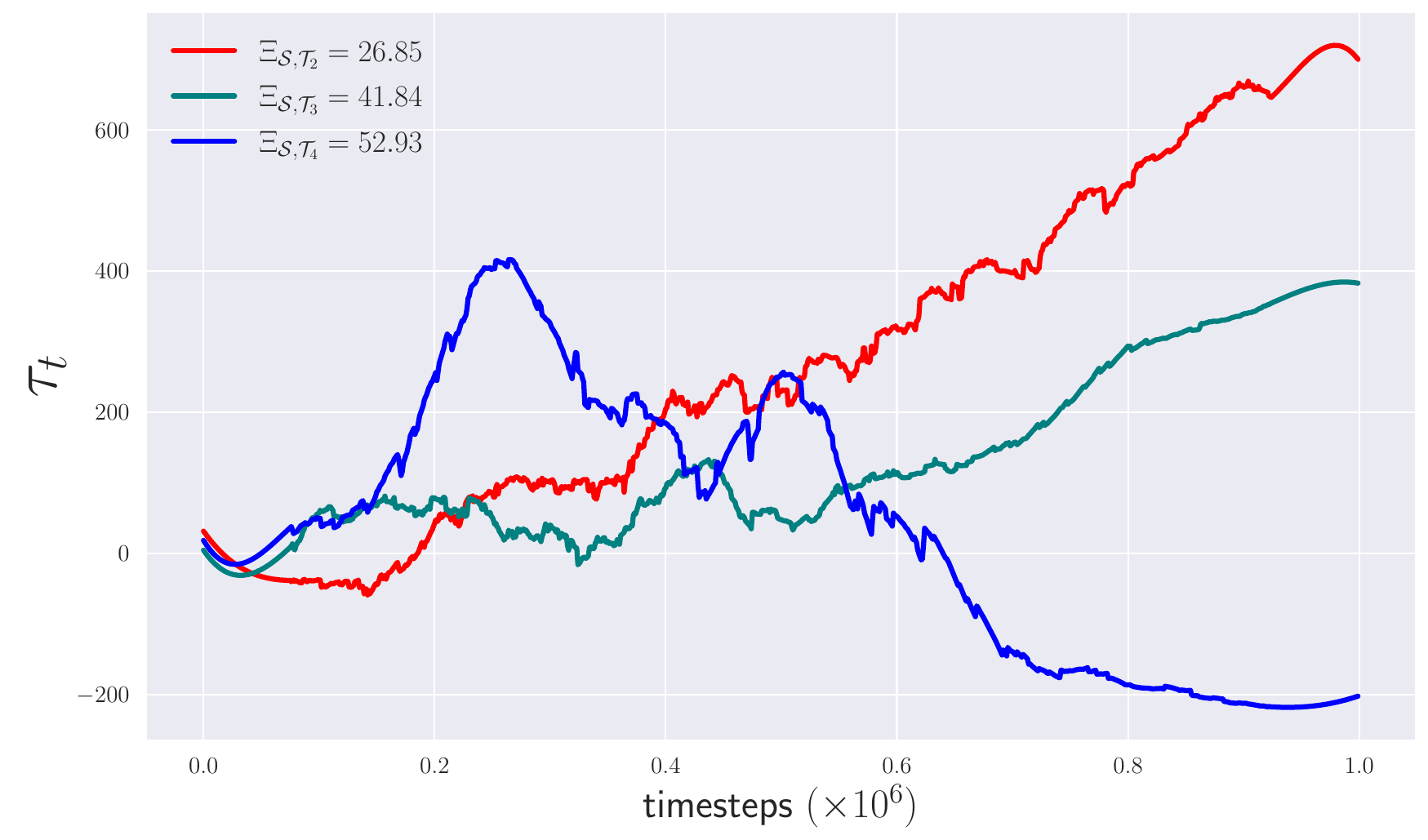} &
   \includegraphics[width=0.40\textwidth]{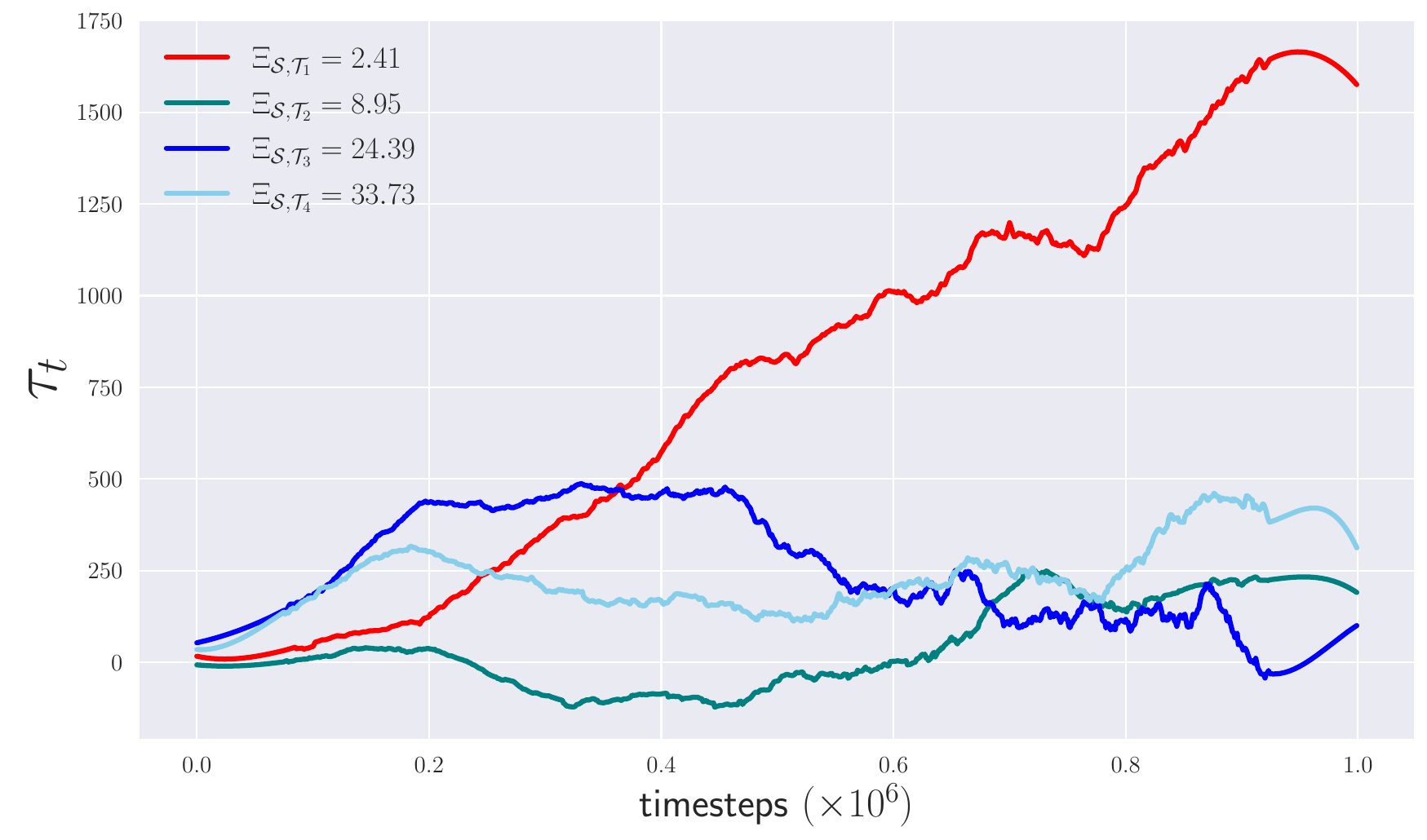} \\
    (a) Half-Cheetah & (b) Ant 
\end{tabular}
\caption{Relative transfer performance for REPAINT against PPO}
\label{fig:tau_repaint}
\end{figure}

\section{Algorithm hyperparameters}\label{app:params}
We keep the hyperparameters the same across all environments. 
\begin{center}
    \begin{tabular}{c c}
    \hline 
     parameter name &  value \\ 
    \hline  
    \begin{tabular}{c} policy network hidden size\\ policy network layers\\learning rate\\replay buffer size\\evaluation steps per epoch\\maximum episode length\\batch size\\ number of gradient updates \end{tabular} & \begin{tabular}{c} 200\\4\\3e-4\\100000\\1000\\1000\\64\\50 \end{tabular}\\
    \hline \\ 
    \end{tabular}
    \label{table:algo_params}
\end{center}

\section{REPAINT algorithm}\label{app:repaint}

The authors utilizes the clipped loss from PPO algorithm\citep{schulman2017proximal} for building the REPAINT algorithm,  
\begin{equation*}
    L_\text{clip}(\theta) = \hat{\mathbb{E}}_t \left[\text{min}\left(l_\theta(s_t, a_t)\cdot\hat{A}_t, \text{clip}_\epsilon(l_\theta(s_t, a_t))\cdot\hat{A}_t \right)\right]  \; \; \text{where} \; \; l_\theta = \frac{\pi_\theta(a_t|s_t)}{\pi_\text{old}(a_t|s_t)}.    
\end{equation*}

A modified version of this clipped loss is expressed using the ratio of target and source policy, 
\begin{equation*}
    L_\text{ins}(\theta) = \hat{\mathbb{E}}_t \left[\text{min}\left(l_\theta(s_t, a_t)\cdot\hat{A}_t, \text{clip}_\epsilon(l_\theta(s_t, a_t))\cdot\hat{A}_t \right)\right] \; \; \text{where} \; \; \rho_\theta = \frac{\pi_\theta(a_t|s_t)}{\pi^*_\mathcal{S}(a_t|s_t)}.    
\end{equation*}

The details of the algorithm are provided in Algorithm \ref{alg:repaint}.

\begin{algorithm}
    \caption{REPAINT algorithm \citep{tao2020repaint}}
    \label{alg:repaint}
    
\begin{algorithmic}[1]
    \State \textbf{Initialize:} $\text{value network parameter} \; \nu,  \text{policy network parameter}\; \theta, \text{source policy} \; \pi_\mathcal{S}^*$
    \State \textbf{Set hyper-parameters:} $\zeta, \alpha_1, \alpha_2, \beta_k$
    \For{$k=1, 2, \dots $}
        \State Set $\theta_\text{old} \leftarrow \theta$
        \State Collect sample $\mathcal{S} = \{(s, a, s', r)\}$
        \State Collect sample $\tilde{\mathcal{S}} = \{(\tilde{s}, \tilde{a}, \tilde{s}', \tilde{r})\}$
        \State Fir state-value network, $V_\nu$ using only $\mathcal{S}$ to update $\nu$
        \State Compute advantage estimates $\hat{A}_1, \dots, \hat{A}_T$ for $\mathcal{S}$ and $\hat{A}'_1, \dots, \hat{A}'_T$ for $\tilde{\mathcal{S}}$
        \For{$t = 1, \dots, T'$}
            \If{$\hat{A}_t < \zeta$} 
                \State Remove $\hat{A}'_t$ and the corresponding transition $(\tilde{s}, \tilde{a}, \tilde{s}', \tilde{r})$ from $\tilde{S}$
            \EndIf
        \EndFor
    \State Compute sample gradient of $L^k_\text{rep}(\theta)$ using $\mathcal{S}$ where
    \begin{equation*}
        L^k_\text{rep}(\theta) = L_\text{clip}(\theta) - \beta_k L_\text{aux}(\theta)
    \end{equation*}
    \State Compute sample gradient of $L_\text{ins}(\theta)$ using $\tilde{\mathcal{S}}$ where 
    \begin{equation*}
        L_\text{ins}(\theta) = \hat{\mathbb{E}}_t [\text{min} \rho_\theta(s_t, a_t)\cdot \hat{A}_t, \text{clip}_\epsilon(\rho_\theta (s_t, a_t))\cdot \hat{A}_t']
    \end{equation*}    
    \State Update policy network by 
    \begin{equation*}
       \theta \leftarrow \theta + \alpha_1 \nabla_\theta L_\text{rep}^k(\theta) + \alpha_2 \nabla_\theta L_\text{ins}(\theta) 
    \end{equation*}    
    
    \EndFor 
\end{algorithmic}
\end{algorithm}

\section{Soft Actor-Critic (SAC) algorithm}\label{app:sac}
Soft actor-critic (SAC) is an off-policy model-free reinforcement learning algorithm. SAC builds upon the maximum entropy objective in RL where the optimal policy aims to maximize both the expected sum of returns and its entropy at each visited state. This can be expressed as the following

\begin{equation}
    \pi^* = \argmax_\pi \sum_t \mathbb{E}_{(\mathbf{s}_t, \mathbf{a}_t)\sim \rho_\pi} [r_t +  \alpha \mathcal{H}(\pi(\cdot|\mathbf{s}_t))],
    \label{eq:max_entropy}
\end{equation}

where $\rho_\pi$ is the state-action marginal of the trajectory distribution induced by the policy $\pi(\cdot|\mathbf{s}_t)$ and $\mathcal{H}(\cdot)$ is the entropy of the policy and $\alpha$ is a temperature parameter to control the effect of entropy. Using this objective it is possible to derive soft-Q values and soft-policy iteration algorithm as the following,

\begin{equation}
    \mathcal{T}^\pi Q(\mathbf{s}_t, \mathbf{a}_t) = r_t + \gamma \mathbf{E}_{\mathbf{s}_{t+1}\sim p}[V(\mathbf{s}_{t+1})],
\end{equation}

where $\mathcal{T}^\pi$ is the modified bellman backup operator and $V(\mathbf{s}_t) = \mathbf{E}_{\mathbf{a}_t\sim \pi}[Q(\mathbf{s}_t, \mathbf{a}_t) - \alpha \log \pi(\mathbf{a}_t|\mathbf{s}_t)].$

Using several approximations to the soft-policy iteration algorithm it is possible to obtain an actor-critic architecture that maximizes the objective in Eq. \ref{eq:max_entropy}. This is specifically done by using deep neural networks to parameterize the value function and the policy. Finally, the objective function can be re-written as the following, 

\begin{equation}
    J(\phi) = \mathbb{E}_{\mathbf{s}_t \sim \mathcal{D}} \left[ D_{\text{KL}}\left( \pi_\phi(\cdot|\mathbf{s}_t) \; \bigg{|}\bigg{|} \; \frac{\exp{(Q_\theta(\mathbf{s}_t, \cdot))}}{Z_\theta(\mathbf{s}_t)}\right) \right]
\end{equation}


\begin{table}[!htb]
    \centering
    \begin{tabular}{c c} \hline 
        Symbol & meaning \\ \hline 
        $\mathcal{M}, \mathcal{M}_\mathcal{S}, \mathcal{M}_\mathcal{T}$ & MDP, source MDP, target MDP\\
        $\mathcal{S}$ & source task\\
        $\mathcal{T}$ & target task\\
        $\mathcal{X}, \mathcal{A}$ & state space, action space\\ 
        $\mathcal{R}_\mathcal{S}, \mathcal{P}_\mathcal{S}$ & source reward, source transition dynamics\\ 
        $\mathcal{R}_\mathcal{T}, \mathcal{P}_\mathcal{T}$ & target reward, target transition dynamics\\ 
        $\mathcal{K}_\mathcal{S}$ & source knowledge\\
        $\mathcal{K}_\mathcal{T}$ & target knowledge\\
        $\mathcal{D}_\mathcal{T}$ & dataset collected in target task\\
        $\mathcal{H}(\cdot, \cdot)$ & cross-entropy\\
        $\pi^*_\mathcal{S}$ & optimal source policy\\
        $\pi_\mathcal{T}$ & target policy\\
        $A_\mathcal{S}, A_\mathcal{T}$ & advantage using source policy, advantage using target policy \\ 
        $\rho$ & transfer evaluation metric\\
        $\tau$ & relative transfer performance\\
        $\Lambda_i$ & transferability of algorithm $i$\\
        $G_t$ & returns at timestep $t$\\
        $\Xi^\mathcal{P}_{\mathcal{S}, \mathcal{T}}$ & dyanmics similarity between $\mathcal{S}, \mathcal{T}$\\        
        $\Xi^\mathcal{R}_{\mathcal{S}, \mathcal{T}}$ & reward similarity between $\mathcal{S}, \mathcal{T}$\\        
        $\beta$ & temperature parameter \\ 
        $\psi$ & source policy parameters e.g. $\pi_\mathcal{S}^* \equiv \pi_\psi$\\
        $\phi$ & target policy parameters e.g. $\pi_\theta$\\
        $\theta$ & value function parameters\\ 
        \hline        
    \end{tabular}
    \caption{Nomenclature}
    \label{tab:nomenclature}
\end{table}

\end{document}